\documentclass[sigconf]{acmart}
\setcopyright{none}
\renewcommand\footnotetextcopyrightpermission[1]{}
\AtBeginDocument{%
  \providecommand\BibTeX{{%
    \normalfont B\kern-0.5em{\scshape i\kern-0.25em b}\kern-0.8em\TeX}}}

\usepackage{style/hadianeh}
\usepackage{style/macros}
\usepackage{style/nick_names}
\usepackage{style/results}
\usepackage[caption=false,font=footnotesize]{subfig}
\usepackage{graphicx}
\usepackage{subeqnarray}
\usepackage{bbm}
\usepackage{relsize}
\usepackage{bbding}
\definecolor{Gray}{gray}{0.9}
\newcommand{\CC}{\cellcolor{Gray}}
\usepackage{tcolorbox}
\usepackage{balance}
\usepackage[keeplastbox]{flushend}

\acmConference[ISCA '22]{The 49th Annual International Symposium on Computer Architecture}{June 18--22, 2022}{New York, NY, USA}
\acmBooktitle{The 49th Annual International Symposium on Computer Architecture (ISCA '22), June 18--22, 2022, New York, NY, USA}

\begin{document}
\title{\texorpdfstring{Accelerating Attention through\\Gradient-Based Learned Runtime Pruning}{Lg}}
\author{%
\texorpdfstring{
Zheng Li$^{\ast}$\quad{}\quad{}Soroush Ghodrati$^{\ast}$\quad{}\quad{}Amir Yazdanbakhsh$^{\ast \dagger}$\vspace{0.05cm}\\
Hadi Esmaeilzadeh\quad{}\quad{}Mingu Kang\vspace{0.2cm}\\
\smaller{
  University of California, San Diego\quad{}\quad{}$^{\dagger}$Google Research \quad{}($^\ast$Equal Contribution) \vspace{0.05cm}}}{Lg}\\
\texorpdfstring{\smaller
{\texttt{zhengli@ucsd.edu, sghodrati@ucsd.edu, ayazdan@google.com, hadi@ucsd.edu, mingu@ucsd.edu}}}{Lg}
}

\renewcommand{\shortauthors}{Li, Ghodrati, Yazdanbakhsh, et al.}
\renewcommand{\shorttitle}{Accelerating Attention through Gradient-based Learned Runtime Pruning}

\begin{abstract}
Self-attention is a key enabler of state-of-art accuracy for various transformer-based Natural Language Processing models.
This attention mechanism calculates a correlation score for each word with respect to the other words in a sentence.
Commonly, only a small subset of words highly correlates with the word under attention, which is only determined at runtime.
As such, a significant amount of computation is inconsequential due to low attention scores and can potentially be pruned.
The main challenge is finding the threshold for the scores below which subsequent computation will be inconsequential.
Although such a threshold is discrete, this paper formulates its search through a soft differentiable regularizer integrated into the loss function of the training.
This formulation piggy backs on the back-propagation training to analytically co-optimize the threshold and the weights simultaneously, striking a formally optimal balance between accuracy and computation pruning.
To best utilize this mathematical innovation, we devise a bit-serial architecture, dubbed \sys, for transformer language models with bit-level early termination microarchitectural mechanism. 
We evaluate our design across 43 back-end tasks for \bench{MemN2N}, \bench{BERT}, \bench{ALBERT}, \bench{GPT-2}, and \bench{Vision} transformer models.
Post-layout results show that, on average, \sys yields \SpeedupOverBaselineAE and \EnergyOverBaselineAE speedup and energy reduction, respectively, while keeping the average accuracy virtually intact ($<\,0.2\%$ degradation). 
\end{abstract}

\maketitle

\section{Introduction}
\label{sec:intro}
Natural Language Processing (NLP) defines the frontier for Artificial Intelligence (AI) as it is even central to the Turing Test~\cite{turingtest}. 
The recent advent of the self-attention mechanism~\cite{vaswani2017attention} enabled unprecedented successes in the field of NLP, shifting the focus of deep learning from convolutional neural networks towards Transformer models in various domains~\cite{jumper2021highly,lineartrans:2020,brown2020language,TrXL:arxiv:2019,gbert:naacl:2019,albert:iclr:2019,t5_model:arxiv:2019,roberta:arxiv:2019,bp-transformer,child2019generating,velivckovic2017graph}.
The self-attention mechanism calculates a score to measure the correlation between a word and all the other words in a subtext.
The subtext is the collection of all the words, which is captured by the attention mechanism.
Therefore, it quantifies the context of the word under attention with respect to its subtext.

Intuitively, a word can bear multiple connotations, of which only one is expressed in its proximate context.
Usually, few keywords define the context and therefore, a significant amount of computation will be inconsequential.
The attention score for a word determines highly correlated words; the rest are merely irrelevant.
There exists a threshold that differentiates between the scores of the words that need to be considered and those that do not define the context and are thus inconsequential.
Because each attention layer identifies a distinct context of the target sentence, such a threshold needs to be defined on a per-layer basis to maintain model accuracy.
Recent research has leveraged this insight and proposed several techniques that skip computation if a threshold is not met~\cite{spatten:hpca21, elsa:isca21,edgebert:micro21,a3:hpca20}.
Clearly, skipping computation negatively impacts model accuracy, which is also dependent on the value of the thresholds.
Therefore, establishing the right thresholds is crucial for the efficacy of the runtime computation pruning methods.
However, the literature~\cite{spatten:hpca21,elsa:isca21,edgebert:micro21,a3:hpca20} has relied on heuristics, statistical sampling, or human input that do not provide reliable expected accuracy.

In contrast, this paper formulates the problem of finding thresholds for the attention layers as a regularizer that amends the loss function of the transformer model.
Our technique is robust even though the threshold values are discrete and cannot be directly optimized through gradient-based approaches.
\emph{A key contribution of this work is to formulate finding the layer-wise pruning thresholds as a differentiable regularizer.
This formulation leverages the gradient-based back-propagation algorithm to mathematically co-optimize the threshold values and weight parameters.}
This approach unblocks simultaneous co-optimization of the two conflicting objectives of maximizing the pruning rate of the computations while minimizing the accuracy loss.
In addition, this analytical technique strikes a formally optimal balance between accuracy and computation pruning.
Note that the current Cambrian explosion of deep learning hinges upon two main algorithmic innovations. 
First, changing the activation function of perceptrons from a non-differentiable step function to the continuous smooth sigmoid function~\cite{backpropagation} enabled back-propagation and multi-layer neural networks and ended the first AI winter~\cite{ai_winter}. 
Second, solving the vanishing gradients problem~\cite{hochreiter2001gradient} has resulted in stable training deep neural networks that have taken the IT industry by storm.
The proposed approach is analytical and therefore mathematically sound, and does not rely on limited empirical evidence.
The solution also guarantees the same generality and optimality  that are essential for training the machine learning model itself.

In this paper, we apply these algorithmic innovations to learn self-attention thresholds in a gradient-based fashion. 
At runtime, the attention scores below the learned threshold are replaced by $-\infty$ to void their impact on the attention layer's outputs.
As such, the preceding computation can be pruned early when the result is below the threshold.
We devise a bit-serial architecture, called \sys\footnote{\textbf{\leopard}: \textbf{L}earning thr\textbf{E}sholds for \textbf{O}n-the-fly \textbf{P}runing \textbf{A}cceleration of t\textbf{R}ansformer mo\textbf{D}els.}, to maximize the benefits by terminating computation even before pruning the following calculation.
This design reduces computation at the finest granularity possible (bit level), hence offering benefits beyond pruning.
Our hardware mechanism for early termination is exact and does not cause any accuracy degradation.
\emph{At the bit level, it can be determined ahead of calculation completions if the partial result of the dot-product can ever exceed the threshold.
Therefore, another advance of this paper is leveraging arithmetic insights for early termination in the microarchitecture without approximation.}

We evaluate the effectiveness of our gradient-based algorithmic innovation and the proposed bit-level arithmetic properties by designing and implementing the \sys accelerator in hardware.
We synthesize and generate layout for a prototype of the \sys accelerator implementation in a 65~nm process technology and characterize its speed and energy consumption under various settings.
We evaluate various state-of-the-art transformer models, including \bench{BERT}, \bench{GPT-2} and \bench{Vision-Transformer}, and datasets forming a benchmark suite of \rev{43} language and vision processing tasks.
On average, the designed accelerator offers \rev{\SpeedupOverBaselineAE} and \rev{\EnergyOverBaselineAE} speedup and energy reduction, respectively, compared to a baseline design without pruning and bit-level early termination support under an iso-area setting.
\sys's notable pruning rate can unlock more benefits, if more chip area budget (15\%) is available.
Given this extra area budget, our accelerator's benefits increase to \SpeedupOverBaselineHP and \EnergyOverBaselineHP speedup and energy reduction, respectively.
To better understand the sources of these improvements, we also distinguish between the effects of runtime computation pruning and bit-level early termination on energy savings.
Our study across the target models shows that, on average, out of the \rev{\EnergyOverBaselineAE} energy reduction, \rev{\PruningEnergy} stems from runtime computation pruning and \rev{\TerminationEnergy} emerges from bit-level early termination.
We also compare \sys to two state-of-the-art accelerators for self-attention mechanism, \aaa~\cite{a3:hpca20} and SpAtten~\cite{spatten:hpca21}, which support runtime pruning.
However, neither accelerator provides analytical support or guarantee for model accuracy, only relying on heuristic approximations.
The results from our evaluations suggest that formulating runtime pruning as a gradient-based optimization can unlock significant benefits, while guaranteeing inference accuracy.
\section{Background and Motivation}
\label{sec:overview}
\subsection{Self-Attention Mechanism}\label{subsec:overview:self-attention}
``\textit{Self-attention}'' is a mechanism to find the relation between a word to all the other words in a sentence~\cite{vaswani2017attention,performer}.
To compute this relation, we first project each word to a vector with $d_w$ dimensions, so-called embedding.
Given a sentence with $s$ words, this projection creates a matrix $\mathcal{X}$ with $s\times{d_w}$.
Then, these word embeddings are multiplied into query weight matrix ($W^\mathcal{Q}$), key weight matrix ($W^\mathcal{K}$), and value weight matrix ($W^\mathcal{V}$), each with $d_w\times{d}$ dimensions as follows:
\begin{equation}
\label{eq:qk}
    \mathcal{Q}_{s\times{d}} = \mathcal{X}\times\mathcal{W^\mathcal{Q}};\quad\mathcal{K}_{s\times{d}} = \mathcal{X}\times\mathcal{W^\mathcal{K}};\quad\mathcal{V}_{s\times{d}} = \mathcal{X}\times\mathcal{W^\mathcal{V}}\\
\end{equation}
\noindent{}Given the query ($\mathcal{Q}$) and key ($\mathcal{K}$) matrices, a self-attention $Score$ matrix is calculated as follows:
\begin{equation}
\label{eq:score}
   \mathrm{Score}_{s\times{s}} = \mathcal{Q}\times\mathcal{K}^T
\end{equation}
\noindent{}where each element $s_{ij}$ in the self-attention $Score$ matrix indicates the relation between word$_i$ and word$_{j}$ in the input sentence.
The $Score$ values are generally scaled down by ($\times 1/\sqrt{d}$) before the next step to enable stable gradients during training~\cite{vaswani2017attention}.
To ensure that the self-attention $Score$s are positive and adding up to one, ``\textit{softmax}'' is applied to each row of $\mathrm{Score}$ matrix as follows:
\begin{equation}
\label{eq:prob}
\begin{split}
  \mathcal{P}_{s\times{s}} &= \mathrm{Softmax}(\mathrm{Score})
\end{split}
\end{equation}
\noindent{}Softmax outputs indicate a $\mathrm{probability}$ estimation of the input words' relation.
The self-attention values are calculated as follows:
\begin{equation}
\label{eq:attention}
   \mathrm{Att}_{s\times{}d} = \mathcal{P}\times\mathcal{V}
\end{equation}

Generally, each attention layer consists of multiple heads each with dedicated $W^\mathcal{Q}$, $W^\mathcal{K}$, and $W^\mathcal{V}$ weight matrices. Each head presumably captures different dependencies between the token embeddings.
In this case, the attention values  (Equation~\ref{eq:attention}) from each head are concatenated and projected into an attention matrix of size $s\times{}d_w$ using a weight matrix $\mathcal{W}^o_{({d}\times{}h)\times{d_w}}$ as follows:
\begin{equation}
\label{eq:multi-attention}
 \mathrm{Multi-Head\;Att}_{s\times{}d_w} = \mathrm{Concat}(\mathrm{Att}_1,\ \mathrm{Att}_2,\ \cdot\cdot\cdot,\ \mathrm{Att}_h)\times\mathcal{W}^o
\end{equation}
\noindent{}where $\mathrm{Concat}$ operation concatenates the $\mathrm{Att}$ output matrix from each head to generate a $(s\times{}(d\times{}h))$-matrix.

\subsection{Gradient-Based Optimization and Regularization}\label{subsec:overview:loss}
\niparagraph{Gradient-based optimization.}
Training neural networks are formulated as an optimization problem of a predefined loss function.
These loss functions are generally non-convex and have a manifold consisting of different local optima which makes the training of neural networks challenging.
To alleviate the complexity of optimizing loss functions, it is common to use gradient-based methods~\cite{robbins1951stochastic,kingma2014adam}. 
Using these gradient-based methods institute defining differentiable loss functions, such as cross-entropy~\cite{murphy2012machine} or Kullback-Leibler divergence~\cite{kullback1951information} which is prevalent in self-attention models~\cite{gbert:naacl:2019,vaswani2017attention,memn2n:nips:2015}.

\niparagraph{Regularization in loss function.}
To impose certain constraints on the model parameters, such as improved generalization~\cite{krogh1992simple,zou2005regularization,srivastava2014dropout} and introducing sparsity~\cite{gale2019state,srinivas2017training,learnedthreshold}, it is common to use regularizer as part of the loss function.
However, using gradient-based methods for training mandates these regularizers to be framed as additional differentiable terms to the loss.
This differentiability constraint for employing gradient-based methods introduces a unique challenge for supporting constraints that are not inherently differentiable.
\subsection{Motivation}
Analyzing the computations for self-attention layers, it is apparent that the main computation cost is associated to $\mathrm{Score}$ (Equation~\ref{eq:score}) and attention computations (Equation~\ref{eq:attention}) that necessitates the multiplications of two matrices with $s\times{}d$ dimensions, each with time complexity of $\mathcal{O}(s^2d)$.
These complexities translate to quadratic raise in computation cost and storage as the number of input tokens increases.
As such, prior work aims to reduce the time and space complexity of these operations both from the algorithmic~\cite{performer,linformer:2020,longformer:2020,bigbird:2020,child2019generating} and hardware perspectives~\cite{spatten:hpca21,elsa:isca21,kao2021optimized,softermax:dac21,a3:hpca20}.
In this work, we propose an alternative pruning mechanism that learns the threshold as part of training.
Our proposed technique prunes away unimportant $Score$s, hence eliminating the ineffectual computations of ``\textit{softmax($\cdot$)}'' in Equation~\ref{eq:prob} and ``$\times\mathcal{V}$'' in Equation~\ref{eq:attention}.
In addition, to further cut down the computations of $Score$s (Equation~\ref{eq:score}), we employ a unique early-compute termination without impacting the model accuracy.
\begin{figure}[t]
\centering
\subfloat[Ideal Pruning]{\label{fig:thresholding_ideal}\includegraphics[width=0.45\linewidth]{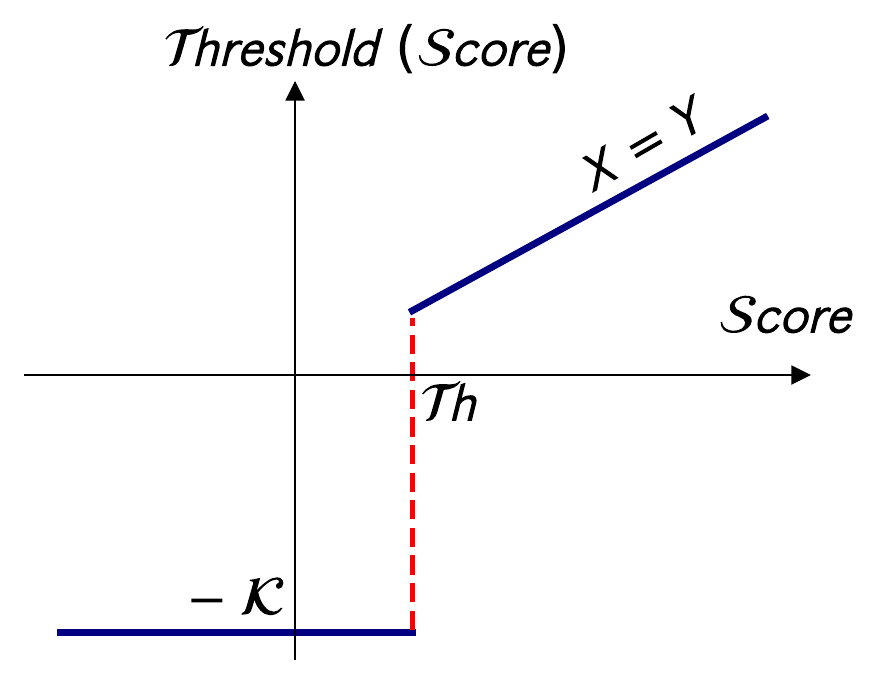}}
\subfloat[Pruning with Soft Threshold]{\label{fig:thresholding_soft}\includegraphics[width=0.45\linewidth]{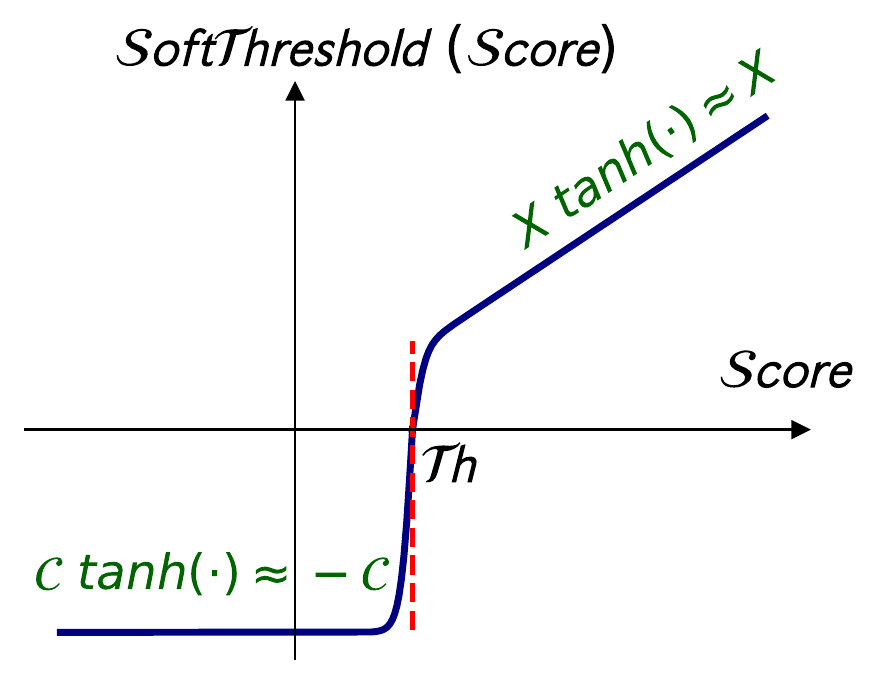}}
\caption{Pruning operation on attention $Score$: (a) ideal magnitude-based pruning, (b) proposed differentiable pruning operation with soft threshold.}
\label{fig:thresholding} 
\end{figure}
\section{Algorithmic Optimizations for Sparse Attention}
\label{sec:algorithm_optimization}
The section overviews the algorithmic optimizations for inducing sparsity in attention layers.
We first introduce an online pruning method that eliminate \textit{unimportant} attention layer computations as early as possible, right after $Score$ calculations (e.g. $\mathcal{Q}\times\mathcal{K}\rev{^T}$), to increase the realized performance benefits.
Particularly, our method sets the layer-wise pruning thresholds as trainable parameters and jointly fine-tune the model parameters and learn the pruning thresholds as part of a light fine-tuning step.
Then, our method compares the $Score$ = $\mathcal{Q}\times\mathcal{K}\rev{^T}$ values against the learned pruning thresholds per attention layer and prunes the ones that satisfy the pruning criteria.
Note that, in contrast to prior learned weight pruning method for image classification models ~\cite{learnedthreshold}, the pruning criteria in our work is content-dependant and is applied adaptively based on the calculated $Score$ values.
That means the induced sparsity in attention layers by our approach varies from one content to another content.
As our results indicate (See Section~\ref{sec:eval}), the adaptive and content-dependant nature of our pruning method enables high sparsity in attention computations while yielding virtually no accuracy loss. 

\subsection{Learned Per-Layer Pruning}
Learning per-layer pruning thresholds for attention layers consists of three main challenges.
First, the search space of threshold values is complex and computationally intractable for exhaustive exploration.
For example, \bench{BERT-L} model has 24 layers creating a total of 24 threshold parameters, each of which can take any continuous value.
Second, simply sweeping the threshold values as a one-time fine-tuning step could negatively affect the model accuracy~\cite{a3:hpca20,spatten:hpca21}.
To mitigate these challenges, we propose to jointly fine-tune the model parameters and learn the threshold values as a light fine-tuning step with the joint objective of increasing model sparsity and retaining the baseline model accuracy.
However, training the threshold values with the inherently non-differentiable pruning operation poses a unique challenge for gradient-based learned methods.
For this, we use an approximate differentiable pruning operation and devise a surrogate regularizer to reinforce sparsity as part of the model loss function.
In the following paragraphs, we expound our learned pruning method that couples two design principles, namely ``\textit{pruning with soft threshold}'' and ``\textit{surrogate L$_0$ regularization}''.

\niparagraph{Pruning with soft threshold.}
Figure~\ref{fig:thresholding_ideal} demonstrates an ideal pruning operation for $Score$ values (e.g. $Score$ = $\mathcal{Q}\times\mathcal{K}\rev{^T}$, where  $\mathcal{Q}$ and $\mathcal{K}$ are $d$-dimension vectors corresponding to a single word).
The $Score$ values greater than $\mathcal{T}h$ remain unchanged and those less than $\mathcal{T}h$ are clipped to a large negative number.
As the pruning operation is followed by a ``\textit{softmax($\cdot$)}'', setting the $Score$ values below $\mathcal{T}h$ to a large negative number makes the output of the softmax operation sufficiently close to zero. Hence, the large negative numbers are pruned out of the following multiplication into $\mathcal{V}$.
However, using this pruning operation as part of a gradient-based training method is not straightforward due to its discontinuity at $X$ = $\mathcal{T}h$.

To circumvent the non-differentiality in the pruning operation, we propose to replace this operation with an approximate function that instead uses a soft threshold (shown in Figure~\ref{fig:thresholding_soft}) as follows:
\begin{equation}
\label{eq:softhreshold}
    \mathrm{SoftThreshold(x)}= 
    \begin{cases}
    {x~\mathrm{tanh}(s(x-\mathcal{T}h))},&  x \geq \mathcal{T}h\\
    {c~\mathrm{tanh}(s(x-\mathcal{T}h))},  &  x < \mathcal{T}h
\end{cases}
\end{equation}
By assigning a reasonably large value to $s$, the shape of $tanh(\cdot)$ around $\mathcal{T}h$ becomes sharper and enables the learning gradients to effectively flow around this region.
Supporting the learning gradients to flow at the vicinity of $\mathcal{T}h$ allow the gradient-based learning algorithm to either push down the model parameters (e.g. $\mathcal{Q}$ and $\mathcal{K}$) below the threshold or lift them above the threshold according to their contributions to the overall model accuracy.

Outside the vicinity of $\mathcal{T}h$, the $tanh(\cdot)$ asymptotically approaches one and the ``SoftThreshold'' function simply becomes $\approx{x}$ and $\approx{-c}$ for values $\geq$ $\mathcal{T}h$ and < $\mathcal{T}h$, respectively, which are close approximations of the original pruning operation.
In our experiments, we empirically find that setting $c$ = 1000 and $s$ = 10 yield a good approximation for pruning and enables robust training.

\niparagraph{Differentiable surrogate \textit{L}$_0$ regularization.}
Using soft threshold as the sole force of pruning does not necessarily increase sparsity.
Intuitively, the training method may just simply lower the threshold to be a small value, which translates to lower sparsity to maintain high model accuracy.
Imposing such constraints to gradient-based methods are generally achieved through adding a regularizer term to the loss function.
A common method to explicitly penalize the number of non-zero model parameters is to use L$_{0}$ regularizer on model parameters in the loss function as follows:
\begin{subeqnarray}
    &&\mathrm{L}_{tot}(\theta) = \frac{1}{N}\Big(\mathlarger{\sum}_{i=1}^{N}\mathcal{L}\big(A(x_{i};~\theta),~y_{i}\big)\Big) + \lambda||\theta||_{0} \\
    &&||\theta||_{0} = \mathlarger{\sum}_{j=1}^{|\theta|} \mathbbm{1}[\theta_{j} \ne 0]
\label{eqn: loss}
\end{subeqnarray}
where $\mathcal{L}$ is the model loss function, $A(\cdot)$ is the model output for given input $x_{i}$ and model parameters $\theta$, $y_i$ is the corresponding labeled data, $\lambda$ is the balancing factor for L$_0$ regularizer, and $\mathbbm{1}$ is the identity operator that counts the number of non-zero parameters.

Similar to ``\textit{Threshold}'' function, L$_0$ regularizer suffers from the same non-differentiability limitation.
To mitigate this, Louizos et al.~\cite{louizos2017learning} uses a reparameterization of model parameters to compute the training gradients. While this reparameterization technique yields state-of-the-art results for Wide Residual Networks~\cite{wrn:2016} and small datasets, a recent study~\cite{gale2019state} shows that this reparameterization trick performs inconsistently for large-scale tasks such as attention models.
In this work, we propose a simple alternative method that uses a differentiable surrogate L$_0$ regularization for the pruning of $Score$ values in attention layers as follows:
\begin{subeqnarray}
\label{eq:sparse_count}
    &&||\theta||_{0} = \mathlarger{\sum}_{j=1}^{|score|} \mathbbm{1}[\mathrm{score}_{j} > -c]\\
\label{eq:sparse_count_diff}
    &&||\theta||_{0} \approx \mathlarger{\sum}_{j=1}^{|score|} \mathrm{sigmoid}({k(\mathrm{score}_{j}+c-\alpha}))
\end{subeqnarray}
where $k$ = 100 and $\alpha$ = 1.
Using these parameters forces the output of sigmoid($\cdot$) to asymptotically approach one for unpruned $Score$ values and zero for the pruned ones, which are already bounded to $-~c$ as shown in Equation~\ref{eq:softhreshold}.
As such, the proposed differentiable surrogate L$_0$ regularizer is a close approximation of the original L$_{0}$ regularizer in Equation~\ref{eq:sparse_count}~(a).
\begin{figure}[t]
\centering
\subfloat[Sparsity and Pruning Threshold]{\label{fig:curve_sparsity_threshold}\includegraphics[width=0.5\linewidth]{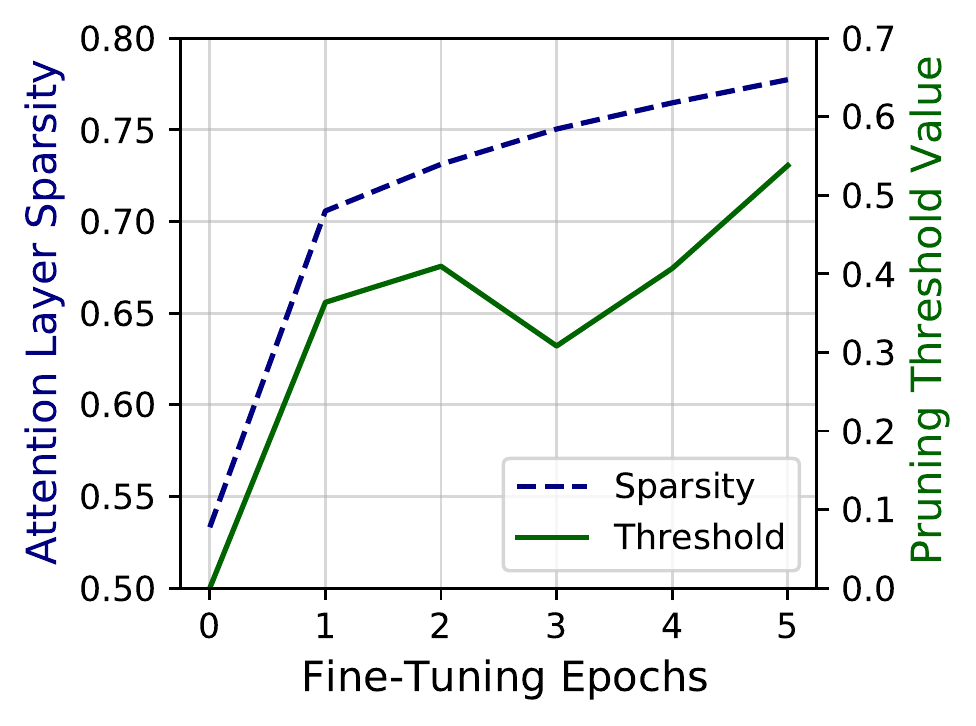}}
\subfloat[Normalized Training Loss]{\label{fig:curve_training_loss}\includegraphics[width=0.43\linewidth]{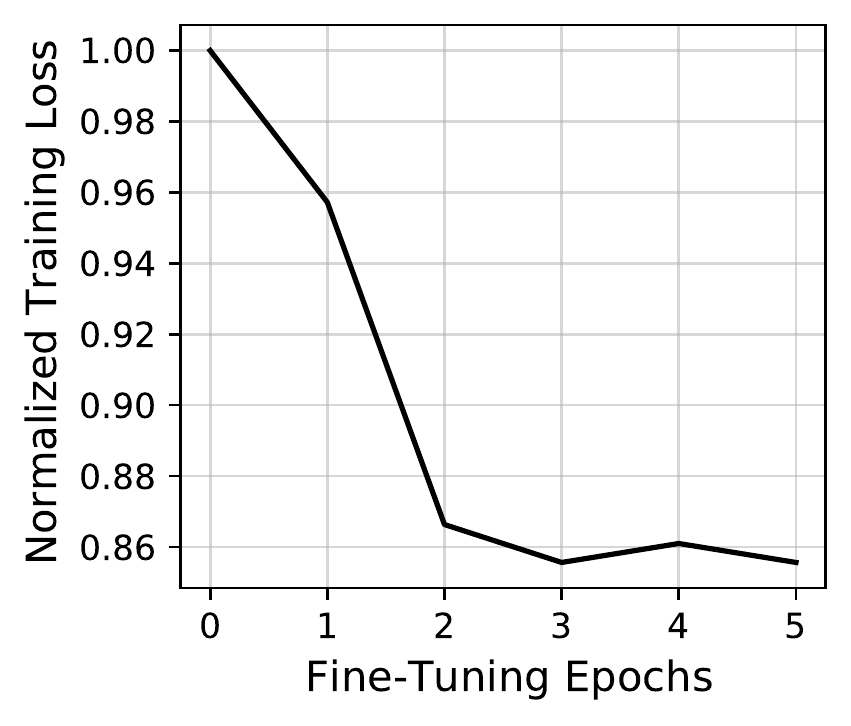}}
\caption{An example (a) attention layer sparsity and its corresponding pruning threshold values and (b) normalized training loss as fine-tuning epochs progress for \bench{BERT-L} model on QNLI task from GLUE benchmark.}
\label{fig:training_curve} 
\end{figure}

\niparagraph{Pruning mechanism.}
We apply our gradient-based learned pruning as a light fine-tuning step based on the previously proposed design principles: (1) pruning with soft threshold and (2) differentiable surrogate \textit{L}$_0$ regularization.
We employ the pre-trained attention models with the proposed modified loss function (e.g. original loss function and the surrogate L$_0$ regularizer) to jointly fine-tune the model parameters and learn the per-layer pruning thresholds.
Using the proposed soft threshold mechanism in the fine-tuning step allows the gradient-based learning method to adjust the model parameters smoothly at the vicinity of the $\mathcal{T}h$ value.
That is, pushing down the non-important model parameters below threshold values and lifting up the important model parameters above it.
One of the main benefits of using the proposed differentiable approach is enabling the model parameters to freely switch between prune and unpruned region.
For all the studied attention models, we initialize the threshold values to zero and run the fine-tuning for up to five epochs.

Figure~\ref{fig:training_curve} demonstrates an example sparsity, threshold values, and normalized training loss curves for \bench{BERT-B} model on QNLI task from the GLUE benchmark.
Figure~\ref{fig:curve_sparsity_threshold} shows that as fine-tuning epochs progress, both the sparsity and threshold values increase owing to the effectiveness of our joint co-training of sparsity and model parameters.
The flexibility afforded by the joint co-training is further illustrated at the third epoch, where the sparsity continues to increase despite the corresponding decrease in the threshold value.
Additionally, Figure~\ref{fig:curve_training_loss} shows the decreasing trend of normalized training loss over the course of fine-tuning epochs.
\subsection{Bit-Level Early-Compute Termination}
\label{sec:early_stop}
The learned pruning offers a unique opportunity to further improve the \sys performance through bit-serial $\mathcal{Q}\times\mathcal{K}\rev{^T}$ computation.
If our system can anticipate that the final result of $\mathcal{Q}\times\mathcal{K}\rev{^T}$ computation is below the learned pruning threshold, the ongoing bit-serial computations can be terminated.
However, this early-termination mechanism poses a key challenge in our design.
As we desire to maintain the baseline model accuracy, the early-termination mechanism must not tamper with the computational correctness of attention layers.
To address this, we propose to compute and add a dynamically adjusted conservative margin value to the partial sum during the bit-serial computations. 
The role of this margin is to account for the maximum potential increase in the remaining $\mathcal{Q}\times\mathcal{K}\rev{^T}$ computations.
If the addition of the partial sum values and the margin still falls below the learned pruning threshold value, the computations are terminated and the corresponding $\mathcal{Q}\times\mathcal{K}\rev{^T}$ is simply pruned.
In the following paragraph, we illustrate the proposed early-compute termination with a conservative margin.

\begin{figure}
\subfloat[Cycle = 1]{\label{fig:step_1}\includegraphics[width=0.245\linewidth]{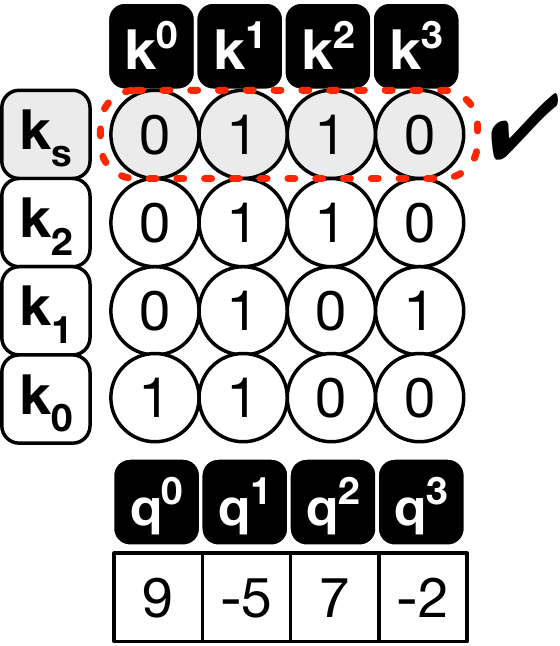}}
\subfloat[Cycle = 2]{\label{fig:step_2}\includegraphics[width=0.245\linewidth]{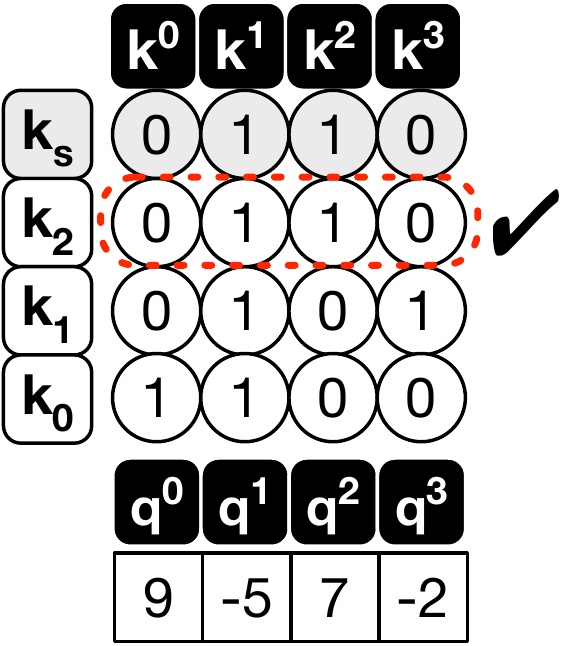}}
\subfloat[Cycle = 3]{\label{fig:step_3}\includegraphics[width=0.245\linewidth]{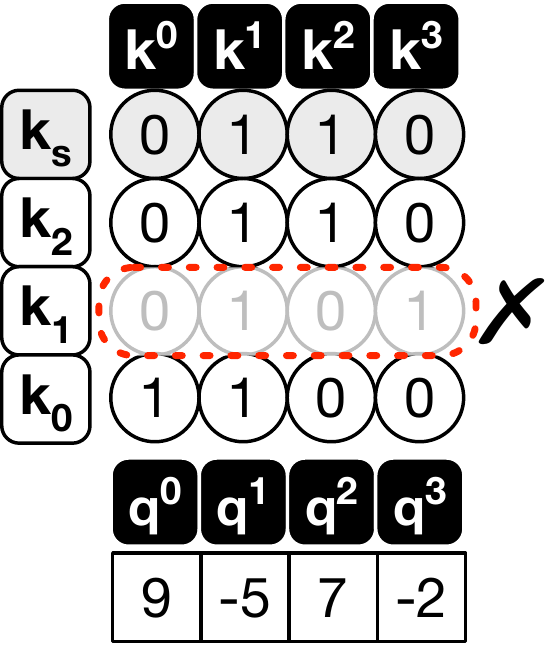}}
\subfloat[Cycle = 4]{\label{fig:step_4}\includegraphics[width=0.245\linewidth]{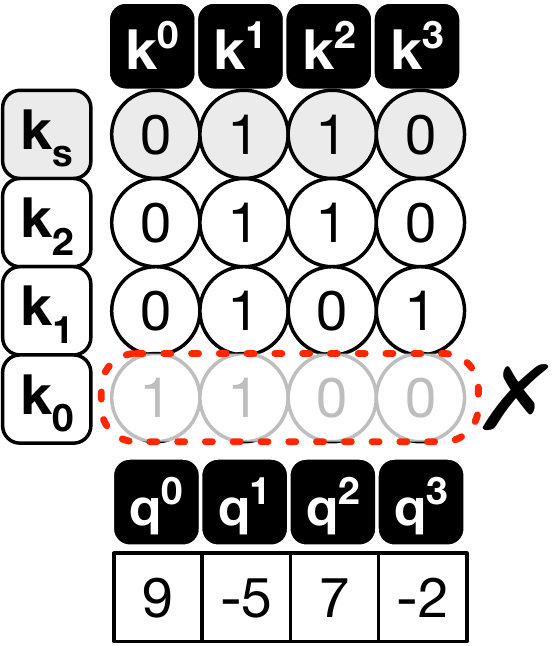}}
\footnotesize{
\centering
\vspace*{0.2cm}
\resizebox{0.475\textwidth}{!}{
\begin{tabular}{l|l|l|l}
\bottomrule
\textbf{Cycle}&\textbf{$\mathcal{P}$ = Partial Sum}&\textbf{$\mathcal{M}$ = Conservative Margin}&\textbf{Early Termination? ($\mathcal{T}h$ = 5)}\\\bottomrule
1&$\mathcal{P}_1 = 0$&$\mathcal{M}_1 = (9+5)(2^{-1}+2^{-2}+2^{-3}) = 12.25$&$\mathcal{P}_1 + \mathcal{M}_1 = 12.25 \geq 5$;~\normalsize{\XSolidBrush}\\\midrule
2&$\mathcal{P}_2 = \mathcal{P}_1 + (5-7)2^{-1} = -1$&$\mathcal{M}_2 = (9+5)(2^{-2}+2^{-3}) = 5.25$&$\mathcal{P}_2 + \mathcal{M}_2 = 4.25 <$ $5$;~\normalsize{\CheckmarkBold}\\\midrule
\CC~3&\CC~$\mathcal{P}_3 = \mathcal{P}_2 + (5-2)2^{-2} = -0.25$&\CC~$\mathcal{M}_3 = (9+5)(2^{-3}) = 1.75$&\CC~$\mathcal{P}_3 + \mathcal{M}_3 = 1.5 < $ $5$;\\\midrule
\CC~4&\CC~$\mathcal{P}_4 = \mathcal{P}_3 + (9+5)2^{-3} = 1.5$&\CC~$\mathcal{M}_4 = 0$&\CC~$\mathcal{P}_4 + \mathcal{M}_4 = 1.5 < $ $5$;\\\bottomrule
\end{tabular}
}
}
\caption{\label{fig:bit_serial_alg_overview}High-level overview of early-compute termination for dot-product operation $\mathcal{Q}\times\mathcal{K}\rev{^T}$. In this example, $\mathcal{K}$ is represented in bit-serial format, whereas $\mathcal{Q}$ is in full-precision fixed-point format. In Figure (a-d) each column illustrate one element of $\mathcal{K}$ vector and each row represents its corresponding bits (MSB $\rightarrow$ LSB). K$_{s}$ indicates the sign bit. For simplicity, $\mathcal{K}$ elements are scaled to be between -1.0 and +1.0. The table shows the partial sum values after each cycle.}
\end{figure}

\niparagraph{Early-compute termination for dot-product operation.}
Figure~\ref{fig:bit_serial_alg_overview} depicts the flow for a $\mathcal{Q}\times\mathcal{K}\rev{^T}$ dot-product computation, each with four elements.
$\mathcal{K}$ elements are placed in bit-serial format vertically from MSB $\rightarrow$ LSB, whereas $\mathcal{Q}$ values are stored in full-precision fixed-point format. In this example, the threshold value is set to five.
For simplicity, we assume the computation is performed in sign-magnitude form, $k_s$ represents the sign-bit for $\mathcal{K}$ vector, and the absolute values of $\mathcal{K}$ elements are less than one.

In the first cycle, the elements with concordant signs, ($k^0$, $q^0$) and ($k^1$, $q^1$), are used for margin initialization.
The intuition here is that \textit{only} the multiplications of elements with concordant signs can contribute positively to the final dot-product result.
Multiplications of elements with opposing signs are ignored to keep the margin conservative and eliminate wrongful early compute terminations.
As shown in the table of Figure~\ref{fig:bit_serial_alg_overview}, both the product of $k^2$ and the $q$ vector as well as the margin are updated.
The margin is adjusted to accommodate the largest possible positive contribution to the final value.
In the second cycle, because the sum of $\mathcal{P}_2$ and $\mathcal{M}_2$ dips below the threshold, the computation process terminates. That is, the subsequent cycles (highlighted in gray) are no longer performed.
Note that, with the proposed margin computation, we ensure that no approximation is introduced in the attention layers. 
Next section discusses the hardware realization for this proposal.
\section{\sys Hardware architecture}
\label{sec:hw_arch}
We design \sys hardware while considering the following requirements based on our algorithmic optimizations:
\begin{enumpackedp}
    \item Leveraging the layer threshold values to detect the unpruned $Score$s and their corresponding indices in the output matrix.
    \item Using bit-serial processing to early-stop the computation of pruned $Score$s and associated memory access .
    \item Processing the $\times \mathcal{V}$ operation for only un-pruned $Score$s to minimize operations while achieving high compute utilization.
\end{enumpackedp}
\subsection{Overall Architecture}
\label{sec:overall_arch}
\begin{figure}[t]
\centering
\includegraphics[width=0.7\columnwidth]{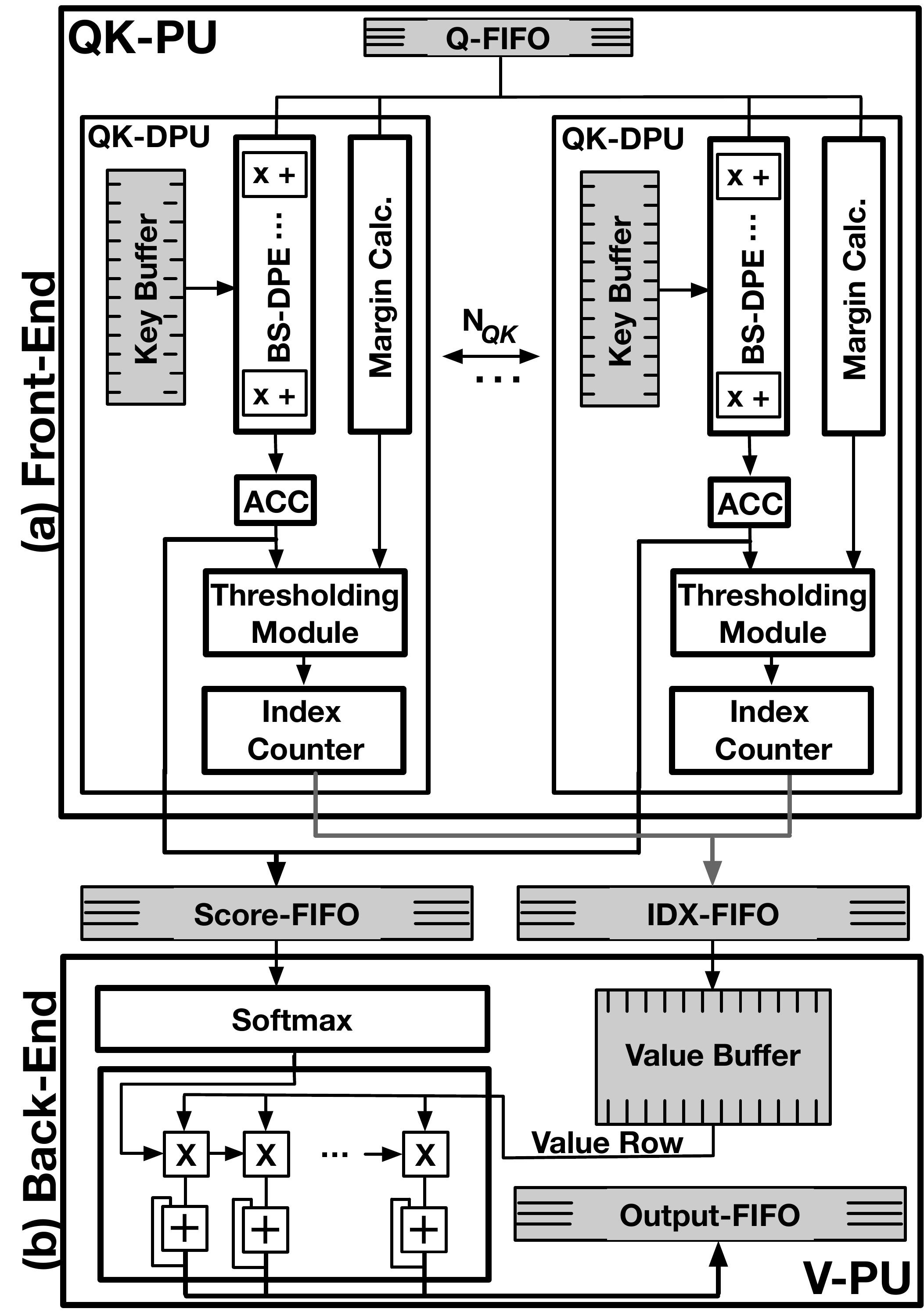}
\caption{Overall microarchitecture of a \sys tile.}
\label{fig:architecture} 
\end{figure}

Due to abundant available parallelism in multi-head attention layers, we design a tile-based architecture for \sys, where attention heads are partitioned across the tiles, and the operations in the tiles are independent of each other on their corresponding heads.
Figure~\ref{fig:architecture} illustrates the high-level microarchitecture of a single \sys tile.
Each tile comprises two major modules to process the computations of attention layers:
\begin{enumpackedp}
\item A front-end unit, dubbed Query Key Processing Unit (\qkvxu), that streams in the $\mathcal{Q}$ vectors (row by row from the $Q$ matrix, where each row corresponds to a word) from the off-chip memory, reads $\mathcal{K}$s from a local buffer, and performs vector-matrix multiplication  between a $\mathcal{Q}$ vector and a $\mathcal{K}$ matrix.
This unit also encompasses a 1-D array of bit-serial dot-product units, \code{QK-DPU}s, each of which equipped with logic to early-stop the computations based on the pruning threshold values and forward the unpruned $Score$s and their indices to the second stage.
\item A back-end unit, dubbed Value Processing Unit (\vvxu), that performs softmax operations on the important un-pruned $Score$s to generate probability, and subsequently performs weighted-summation of the $\mathcal{V}$ vectors  read from a local buffer to generate the final output of the attention layer.
\end{enumpackedp}
The front- and back-end stages are connected to each other through a set of FIFOs that store the survived $Score$s and their corresponding indices. 
The front-end unit employs multiple ($N_{\mathrm{QK}}$) \code{QK-DPU}s while sharing the single \vvxu in consideration of high pruning rate during the processing in the front-end stage.
If the front-end finished the computation with current $\mathcal{Q}$ vector, but the back-end is still working on the previous $\mathcal{Q}$ vector, the front-end unit is stalled until the completion of back-end unit.
As the choice of $N_{\mathrm{QK}}$ is a key factor to maximize the  overall throughput and back-end resource utilization, we explore this design space in Section~\ref{sec:design_space}, which leads to two choices of $N_{\mathrm{QK}}=$ 6 and 8 by focusing on area efficiency and higher utilization, respectively.
Before the operation begins, all the $\mathcal{K}$ and $\mathcal{V}$ matrices are fetched from off-chip memory and stored on on-chip buffers, while the $\mathcal{Q}$ vectors are steamed in.
Since the vectors are re-used by the number of sequence elements (e.g., 512 in BERT), DRAM costs are amortized.
\subsection{Online Pruning Hardware Realization via Bit-serial Execution}
As discussed in Section~\ref{sec:algorithm_optimization}, to realize the pruning of redundant $Score$s during runtime and even earlier termination  with bit-level granularity, we design \sys front-end unit (depicted in Figure~\ref{fig:architecture}-(a)) as a collection of bit-serial dot-product units (\code{QK-DPU}).

\niparagraph{Overall front-end execution flow.}
To perform the $Score$ computations, the $\mathcal{Q}$ vectors are read sequentially from \code{Q-FIFO} and then broadcasted to each \code{QK-DPU}, while each \code{QK-DPU} reads a $\mathcal{K}$ vector from its local \code{Key Buffer} and performs a vector dot-product operation.
As such, while the $\mathcal{Q}$ vector is shared amongst the \code{QK-DPU}s, the $\mathcal{K}$ matrix is partitioned along its columns and is distributed across the \code{Key Buffer}s.
Each \code{QK-DPU} performs the dot-product operations in a \emph{bit-serial} mode, where the $\mathcal{K}$ elements are processed in bit-sequential manner and the $\mathcal{Q}$ elements are processed as a whole (e.g. 12 bit).
Whenever each \code{QK-DPU} finishes the processing of all its $\mathcal{K}$ bits for unpruned $Score$s or early terminates the computation due to not meeting the layer pruning threshold based on the margin calculation described in Section~\ref{sec:early_stop}, it proceeds with the execution of next $\mathcal{K}$ vector.
If a \code{QK-DPU} detects a unpruned $Score$, it stores the $Score$ value and its corresponding index on \code{Score-FIFO} and \code{IDX-FIFO}, respectively, to be processed by the back-end unit later.
Once all the \code{QK-DPU}s finish processing all their $\mathcal{K}$ vectors, the \code{QK-PU} reads the next $\mathcal{Q}$ vector from \code{Q-FIFO} and starts its processing.

\begin{figure}
\centering
\includegraphics[width=0.96\columnwidth]{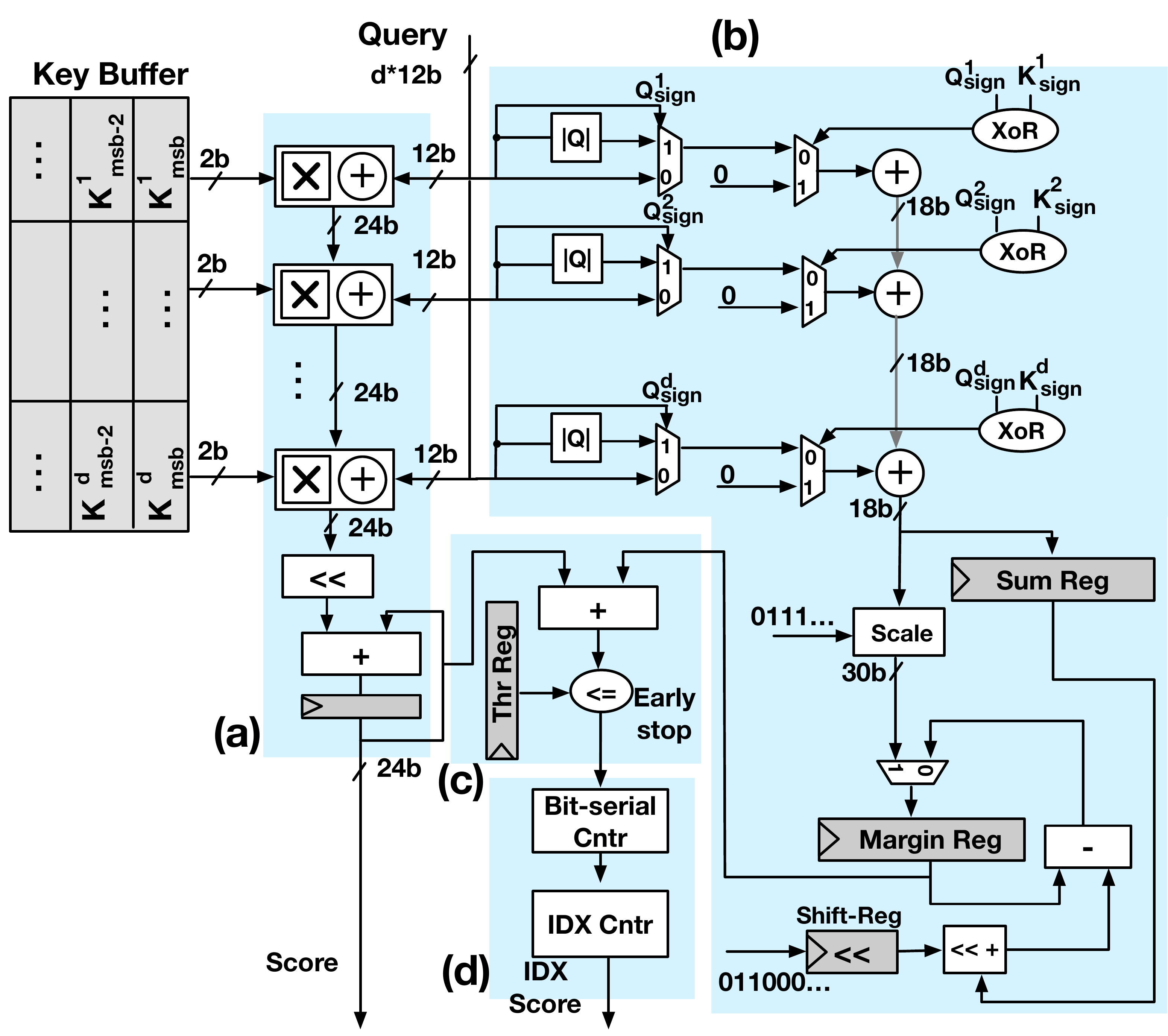}
\caption{A QK-DPU comprising (a) bit-serial dot-product engine, (b) margin calculation logic, (c) thresholding module, and (d) score index counter.}
\label{fig:qk-dpu}
\end{figure}

\niparagraph{Bit-serial dot-product execution.}
Figure~\ref{fig:qk-dpu}-(a) depicts the microarchitectural details of our Bit-Serial Dot-product Engine (\code{BS-DPE}).
The \code{BS-DPE} is a collection of Multiply-ACcumulate (MAC) units and it performs a 12-bit$\times \mathcal{B}$-bit dot-product operation per cycle, where the $\mathcal{Q}$ vector is kept in a local register and $\mathcal{K}$s are read from the \code{Key Buffer} $\mathcal{B}$-bit at a time in a sequential mode.
We chose $\mathcal{B}=2$-bit  as opposed to conventional bit-by-bit serial designs as the number of bits processed per cycle opens a unique trade-off space for the design of \sys.
Increasing the bits leads to better power efficiency due to less frequent latching of intermediate results, however it may degrade the performance as it reduces the resolution of  bit-level early  termination.
As such we perform a design space exploration (Figure~\ref{fig:bit_vs_power} in Section~\ref{sec:design_space}) and chose 2-bit serial execution as it strikes the right balance between power efficiency and performance.
The \code{BS-DPE} accumulates all the intermediate results in around 20 bits to keep required precision of the computations.
The output of the last 2-bit$\times$12-bit MAC unit then goes to a shifter to scale the partial results according to the current $\mathcal{K}$ bit position and is accumulated and stored in a register that holds the (partial) results of $Score$ computations.

\niparagraph{Pruning detection via dynamic margin calculation.}
As discussed in Section~\ref{sec:early_stop} and Figure~\ref{fig:bit_serial_alg_overview}, to detect whether a current $Score$ needs to be pruned and corresponding computations be terminated, \code{QK-DPU} dynamically calculates a \emph{conservative upper-bound margin} ($\mathcal{M}$) and adds it with the current dot-product partial sum ($\mathcal{P}$) to compare it with the layer threshold ($\mathcal{T}h$).
Figure~\ref{fig:qk-dpu}-(b) and (c) show the details of hardware realization for margin calculation and thresholding logic, respectively.
To calculate the margin according to Table in Figure~\ref{fig:bit_serial_alg_overview}, the margin calculation module first detects the $\mathcal{Q}$ and $\mathcal{K}$ pairs in the dot-product that yield positive product.
To do so, during the processing of $\mathcal{K}$'s MSBs, the sign bits of $\mathcal{Q}$s and $\mathcal{K}$s are XORed.
Only if the result is positive (XOR = 0),  the absolute values of the corresponding $\mathcal{Q}$ are summed up to calculate the margin (e.g., resulting in $(9+5)$ in the Table of Figure~\ref{fig:bit_serial_alg_overview}).
The summation result is stored in a  \code{Sum Register}. Then, it is  scaled by the fixed number, largest positive value (e.g. 0111...), which corresponds to  $(2^{-1}+2^{-2}+2^{-3} + ...)$ in Figure~\ref{fig:bit_serial_alg_overview}, storing $(9+5)(2^{-1}+2^{-2}+2^{-3} + ...)$ in the margin register.
%
%
The margin needs to be calculated dynamically for each bit position during bit-serial execution (such as $\mathcal{M}$ changing in each row of the Table in Figure~\ref{fig:bit_serial_alg_overview}).
This is enabled by subtracting the shifted version of \code{Sum Register} value  from the current margin in the margin register, e.g., $(9+5)(2^{-1}+2^{-2}+2^{-3} + ...) - (9+5)(2^{-1}) = (9+5)(2^{-2}+2^{-3} + ...)$ in the second row of the Table in Figure~\ref{fig:bit_serial_alg_overview}.
This operation is iterated every bit position to generate the values in the subsequent rows of the Table in Figure~\ref{fig:bit_serial_alg_overview}.
Note that, the margin calculation is a scalar computation (mostly shift and subtraction), which is amortized over the $d=64$ dimension vector processing, incurring virtually no overhead.
%
After each cycle of the bit-serial operation, the thresholding module (Figure~\ref{fig:qk-dpu}-(c)) adds the updated partial sum with the current margin and compares it with the layer threshold $\mathcal{T}h$ to determine the continuation of the dot-product or its termination for pruning of the current $Score$.

\niparagraph{Final score index calculation.}
The \code{QK-DPU} calculates the indices of the unpruned $Score$s using a set of two counters, as shown in Figure~\ref{fig:qk-dpu}-(d).
First, \code{Bit-serial Cntr} increments with the number of bits processed by the \code{QK-DPU} and gets reset whenever it reaches its maximum (i.e. 6 (= 12bit/$\mathcal{B}$)) for processing all bits for unpruned $Score$s) or the \code{Early stop} flag is asserted.
Second, the value of \code{IDX Cntr} shows the position of the current $Score$ in the vector and increments whenever the \code{Bit-serial Cntr} gets reset, ending the computation of that $Score$.
Finally, if the \code{IDX Cntr} increments and the \code{Early stop} flag is low, the \code{QK-DPU} pushes the content of this counter to \code{IDX FIFO}, because it means that the corresponding $Score$ is not pruned and will be used for further processing in the \code{V-PU}.
\subsection{Back-End Value Processing}
As shown in Figure~\ref{fig:architecture}-(b), the \sys tile's back-end stage, \code{V-PU}, consumes the unpruned $Score$s and executes the Softmax operation, followed by multiplication with $\mathcal{V}$ vectors and finally storing the results to an \code{Output-FIFO}.
Whenever the \code{Score-FIFO} is not empty, the \code{V-PU} starts the Softmax operation ($e^x$ and accumulation) to calculate the probabilities.
We implemented the Softmax module of \code{V-PU} similarly to the Look-Up-Table (LUT)-based methodology in \aaa~\cite{a3:hpca20}.
Whenever the output probability is produced, the \code{V-PU} uses the indices of the unpruned $Score$s to read the corresponding $\mathcal{V}$ vector.
Finally, the $\mathcal{V}$ vector is weighted by the output of the Softmax module with a 1-D array of MAC units.
The elements of $\mathcal{V}$ vector are distributed and the probabilities are shared across the MAC units, similar to a 1-D systolic array.
With such design, the \code{V-PU} consumes the $Score$s sequentially to complete the weighted-sum of $\mathcal{V}$ vectors, and accumulates the partial results over multiple cycles while only accessing the unpruned $\mathcal{V}$ vectors.
As such, it rightfully leverages the provided pruning by the front-end stage and eliminates the inconsequential computations.
\section{Evaluation}
\label{sec:eval}

\subsection{Methodology}
\label{subsec:methodology}

\niparagraph{Workloads.}
We evaluate \sys on various NLP and Vision models: \bench{BERT-Base} (\bench{BERT-B})~\cite{gbert:naacl:2019}, \bench{BERT-Large} (\bench{BERT-L})~\cite{gbert:naacl:2019}, \bench{MemN2N}~\cite{memn2n:nips:2015}, \bench{ALBERT-XX-Large}(\bench{ALBERT-XX-L})~\cite{albert:iclr:2019}, \bench{GPT-2-Large} (GPT-2-L)~\cite{gpt2}, and \bench{ViT-Base}(\bench{ViT-B})~\cite{vit}.
To evaluate these models, we use five different datasets: (1) Facebook \bench{bAbI}, which includes 20 different tasks~\cite{babi:arxiv:2015} for \bench{MemN2N}, (2) General Language Understanding Evaluation (\bench{GLUE}) with nine different tasks~\cite{glue:arxiv:2018} for \bench{BERT} models,  (3) Stanford Question Answering Dataset (\bench{SQUAD}) \cite{squad:arxiv:2016} with a single task for \bench{BERT} models
and \bench{ALBERT-XX-L}, 
(4) \bench{WikiText-2}~\cite{wikitext2} for \bench{GPT-2-L},
and (5) \bench{CIFAR-10}~\cite{cifar10} for \bench{ViT}.
The dimension ($d$) of $\mathcal{Q}$, $\mathcal{K}$, and $\mathcal{V}$ vectors for all the workloads is 64 except \bench{MemN2N} with \bench{bAbI} dataset, which is 20.
The sequence length is 50 for \bench{MemN2N} with \bench{bAbI} whereas 512 and 384 for \bench{BERT} and \rev{\bench{ALBERT-XX-L}} models with \bench{GLUE} and \bench{SQUAD} datasets, respectively.
\rev{Finally, the sequence length for \bench{GPT-2} with \bench{WiKiText-2} is 1280}.
\begin{table}[t!]
\footnotesize
\centering
\caption{Microarchitectural configurations of a \sys tile.}
\resizebox{0.47\textwidth}{!}{
\begin{tabular}{l|l}
\toprule

\textbf{Hardware modules} & \textbf{Configurations} \\\toprule
QK-PU & 6 / 8 {\footnotesize{\code{QK-DPU}}}  (=$N_{\mathrm{QK}}$), each 64 (=$\mathcal{D}$) tap 12$\times$2 bit-serial  \\\hline
Key Buffer& 48KB in total (= 8KB$\times$6 / 6KB$\times$8 banks), 128-bit port per bank \\\hline
V-PU & Single 1-D 64 (=$\mathcal{D}$) way 16$\times$16-bit MAC array\\\hline
Value Buffer & 64KB (= 8KB $\times$ 8 banks), 128-bit port per bank \\\hline
Softmax & 24-bit input,  16-bit output, LUT: 1 KB\\\hline
Score and IDX FIFOs & 24-bit $\times$ 512 depth for Score, 8-bit $\times$ 512 depth for IDX\\
\bottomrule
\end{tabular}
}
\label{table:arch_config}
\end{table}

\niparagraph{Fine-tuning details.}
We use the baseline model checkpoints from HuggingFace~\cite{huggingface:2019} with PyTorch v1.10~\cite{pytorch} and fine-tune the models on an Nvidia RTX 3090, except for \bench{GPT-2-Large}, for which we use an Nvidia A100.
For default task-level training, we use the Adam optimizer with default parameters and the learning rate of $[2, 3]\times{}e^{-5}$ (same as baseline).
To obtain the layer-specific threshold values, we perform an additional pruning-aware fine-tuning step for one to five more epochs to learn the optimal values while maintaining the baseline model accuracy.
For this step, we use the learning rate of $1e^{-2}$ for $\mathcal{T}h$ ($5e^{-6}$ for the other parameters), as training for the $\mathcal{T}h$ is generally slower and a higher learning rate facilitates convergence.
To leverage faster fixed-point execution, we perform a final post-training quantization step with 12 bits for inputs in \code{QK-PU} hardware block and 16 bits for \code{V-PU} block similarly to \cite{spatten:hpca21}. 

\niparagraph{Hardware design details.}
Table~\ref{table:arch_config} lists the microarchitectural parameters of a single \sys tile for two studied configurations:
(1) A \sys tile with six and (2) eight \code{QK-DPU}s that share a single 1-D MAC array in \code{V-PU}.
The number of \code{QK-DPU}s is set such that the compute utilization for front-end and back-end units is balanced, while considering the pruning and bit-level early-termination rates across all the workloads.
We synthesised and performed Placement-and-Route (P\&R) for our designs with two tiles.
The on-chip memory sizes for $\mathcal{K}$ and $\mathcal{V}$ are designed to store up to 512 sequences for a single head in a layer for both configurations.

\niparagraph{Accelerator synthesis and simulations.}
We use Cadence Genus 19.1~\cite{genus} and Cadence Innovus 19.1~\cite{innovus} to perform logic synthesis, floorplan, and P\&R  for the \sys accelerator.
We use TSMC 65~nm GP (General Purpose) standard cell library for the synthesis and layout generation of the digital logic blocks.
These digital blocks are rigorously generated to meet the target frequency of 800MHz in consideration of  all the CMOS corner variations and temperature conditions from $-40^{\circ}$ to $125^{\circ} \mathrm{C}$.
For the SRAM on-chip memory blocks, we use  Memory Compiler with ARM High density 65~nm GP 6-transistor based single-port SRAM version r0p0 \cite{mem_compiler}.

We also develop a simulator to obtain the total cycle counts and number of accesses to memories for both \sys and baseline accelerators.
The simulator incorporates the pruning rate and the bit-level early-termination statistics for each individual workload.
Using these statistics, the simulator evaluates runtime and total energy consumption of the accelerators.

\niparagraph{Comparison to baseline architecture.}
We compare \sys to a conventional baseline design without any of our optimizations (e.g. runtime pruning and bit-level early compute termination).
For a fair comparison, we use the same frequency, bitwidths for $\mathcal{Q} \times \mathcal{K}\rev{^T}$ and $\times \mathcal{V}$, and on-chip memory capacity for all the designs.
The baseline design employs a single 12$\times$12-bit \code{QK-DPU} as opposed to multiple 12$\times$2-bit-serial ones, while both designs have the same back-end \code{V-PU}.
As shown in Table~\ref{table:arch_config}, we evaluate \sys under two design configurations.
The first design with six \code{QK-DPU}s, dubbed Area-Efficient \sys (\aeleopard), almost perfectly matches the area of the baseline design (< 0.2\% overhead) and provides an iso-area comparison setting.
The second one with eight \code{QK-DPU}s, dubbed Highly-Parallel \sys (\hpleopard), provides an area 15\% larger than baseline and delivers a better balance in the compute utilization of the front-end and back-end stages. 

\niparagraph{Comparison with \aaa and SpAtten.}
We also compare \sys with two state-of-the-art attention accelerators, \aaa~\cite{a3:hpca20} and SpAtten~\cite{spatten:hpca21}, with support for runtime pruning.
\aaa employs token pruning by comparing the Softmax output ($\mathrm{probability}$) to a relative threshold, which is set using a user-defined parameter that adjusts the level of approximation.
\aaa also employs a sorting mechanism to make the pruning decision after processing only a small number of large elements from the sorted $\mathcal{K}$ matrix in the order of magnitude.
SpAtten performs cascaded head and token pruning by comparing the Softmax output with a user-defined threshold obtained empirically.
There are no raw performance/energy results for individual workloads and simulation infrastructures of the accelerators. Therefore, we follow the comparison methodology of SpAtten~\cite{spatten:hpca21}, using throughput (GOPs~/~s), energy efficiency (GOPs~/~J), and area efficiency (GOPs~/~s~/~mm$^2$) metrics to provide the best comparisons.
Both \aaa and SpAtten are implemented in 40~nm technology. 
To provide a fair comparison, we scale \hpleopard from 65~nm to 40~nm \rev{based on both Dennard scaling (indicated with $^{\dagger}$) and measurement-based scaling rules \cite{scale_measure} (indicated with $^{\ddagger}$)}.
We use a single tile with an area comparable to \aaa and SpAtten.
Moreover, \aaa implements the $\mathcal{Q} \times \mathcal{K}\rev{^T}$ using 9 bits as opposed to 12 bits in \sys.
As such, we scale the \code{QK-PU} of \hpleopard from 12 bits to 9  bits to provide a head-to-head comparison with \aaa.
\subsection{Accuracy and Algorithmic Optimization} 
\label{subsec:acc_results}
\begin{figure}
\centering
\includegraphics[width=0.97\columnwidth]{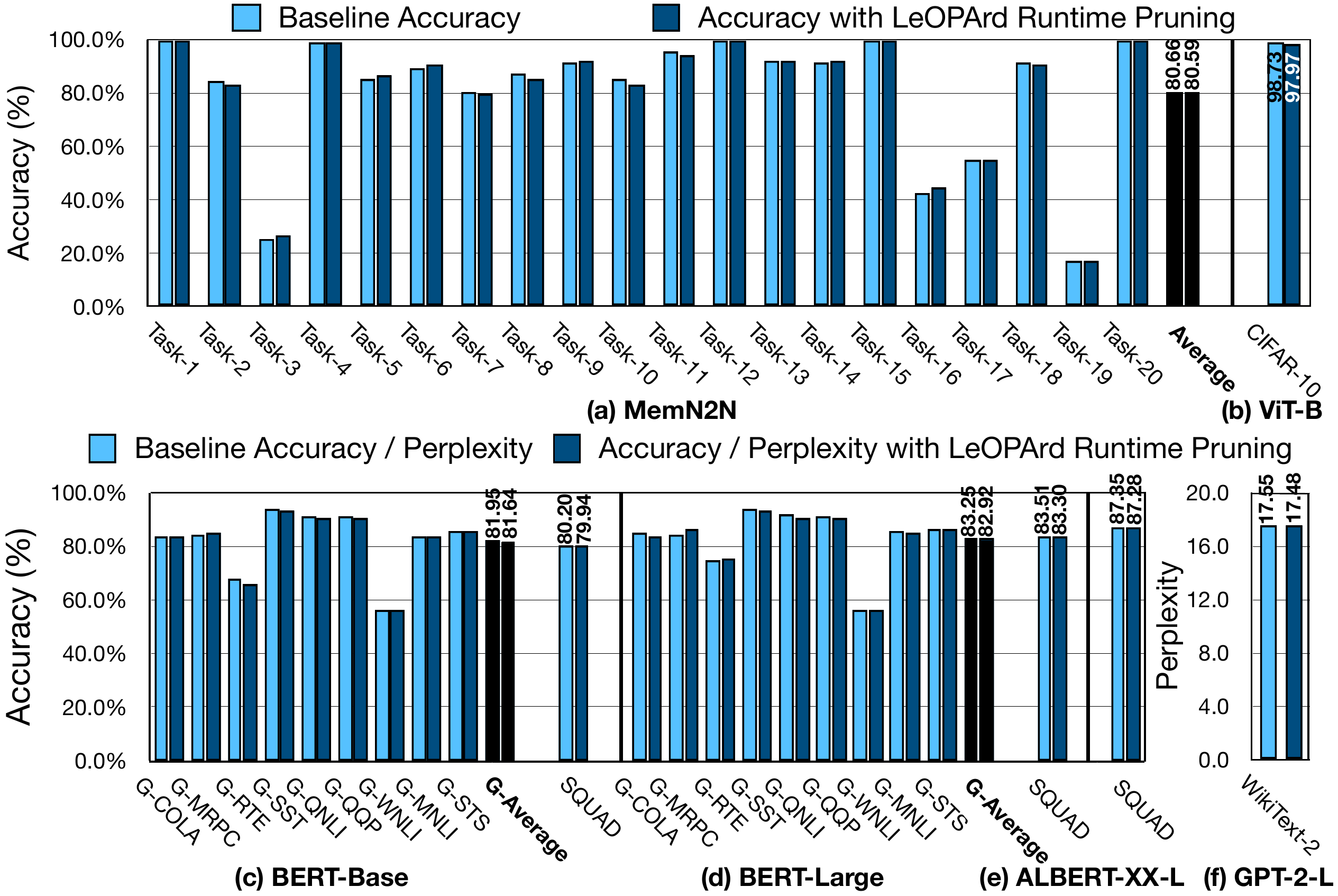}
\caption{Accuracy before and after pruning-aware fine-tuning (prefix "G-": GLUE). We evaluate GPT-2 using perplexity, which favors a lower value.}
\label{fig: accuracy} 
\end{figure}
\niparagraph{Impacts on model accuracy.}
Figure~\ref{fig: accuracy} compares the accuracies of the \sys gradient-based on-the-fly pruning method and the baseline models in their vanilla implementation~\cite{huggingface:2019}, across various tasks of evaluated workloads.
On average, across all the evaluated tasks, \sys runtime pruning degrades accuracy by only 0.07\% for \bench{MemN2N} with the \bench{bAbi} dataset, 0.31\% and 0.33\% for \bench{BERT-B} and \bench{BERT-L} with the \bench{GLUE} dataset, and 0.26\% and  0.21\% for \bench{BERT-B} and \bench{BERT-L} with the \bench{SQUAD} dataset.
For \bench{ALBERT-XX-L} with the \bench{SQUAD} dataset, the \sys runtime pruning leads to only an 0.07\% accuracy loss, whereas the degradation for \bench{ViT-B} with the \bench{CIFAR-10} dataset is 0.76\%.

In the \bench{GPT-2-L} model, we use perplexity, which is the key metric for auto regressive language models.
Note that perplexity is derived from the model loss, and thus lower perplexity is better.
As shown in Figure~\ref{fig: accuracy}-(f), \sys runtime pruning results in a 0.07 decrease in perplexity.
This is achievable because \sys \emph{learns} the optimal threshold values and co-adjusts them with the weight parameters simultaneously via gradient-based optimization.
Figure~\ref{fig: accuracy} also illustrates that the \sys pruning-aware fine-tuning pass evenly improves the accuracy for some of the benchmark tasks, with the maximum of 2.2\%. 
However, this also degrades the accuracy for other tasks with the maximum of 2.6\%.
This accuracy fluctuations are unavoidable due to randomness in deep learning training, but overall the accuracy degradation, averaged across the evaluated benchmarks, converges adequately to a near-zero value ($\leq$\,0.2\%).
Performing the post-training quantization adds at most only 0.1\%, for both the baseline and our pruning-aware fine-tuned models.
\begin{figure}
\centering
\includegraphics[width=0.98\columnwidth]{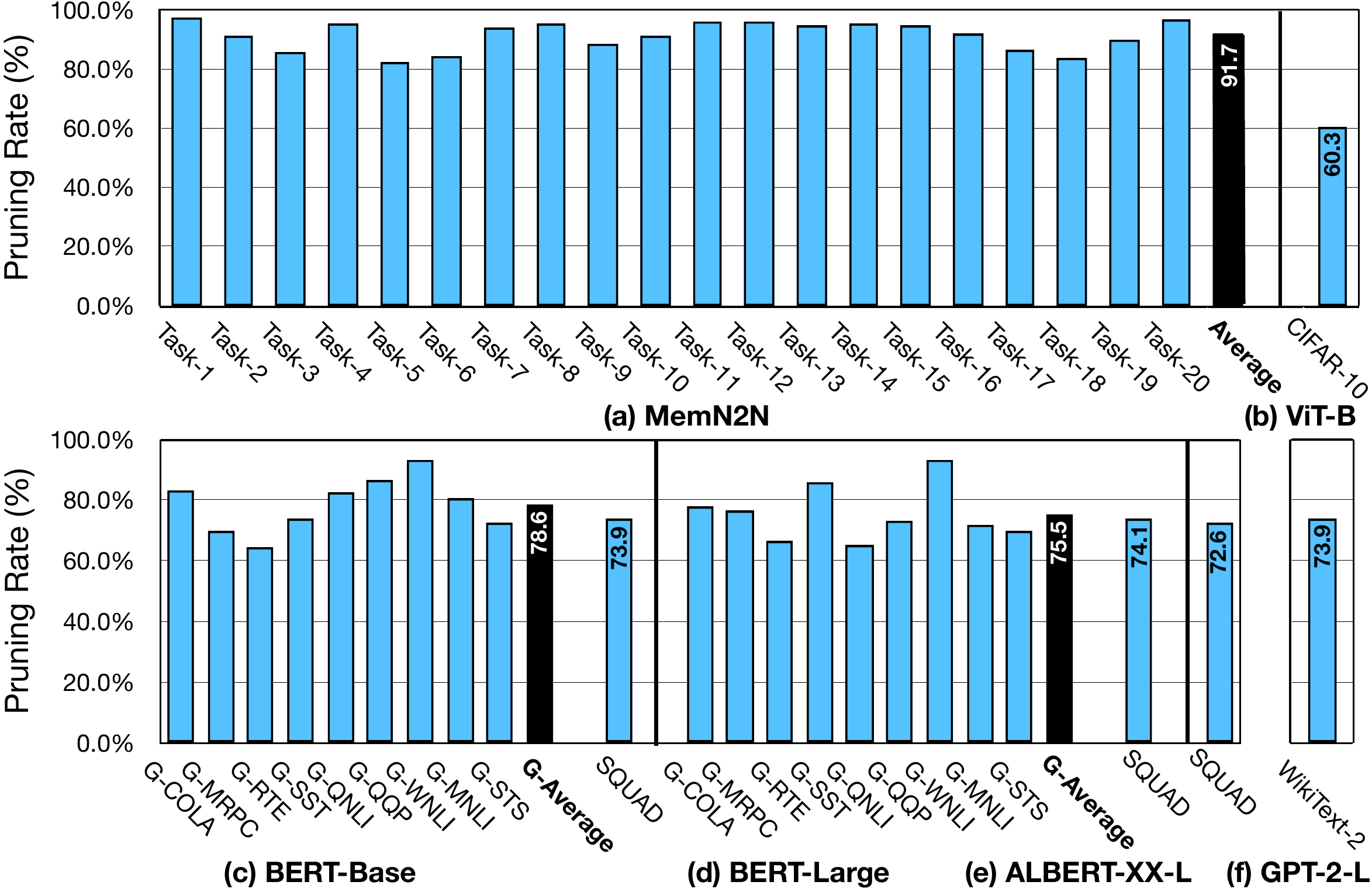}
\caption{Runtime pruning rate with \sys. (prefix "G-": GLUE)}
\label{fig:sparsity}
\end{figure}

\niparagraph{Runtime pruning rate analysis.}
Figure~\ref{fig:sparsity} shows the percentage of total $\mathcal{Q}\times\mathcal{K}\rev{^T}$ $Score$s that are pruned away by our method using the learned threshold values across various benchmarks.
In transformer software implementations, zeros are padded to maintain regular vector length despite the varying sequence length in each workload. The padded zeros are not counted for sparsity contribution in this paper.
On average, \sys prunes 91.7\% (max. 97.4\%) of $Score$s  across all the 20 tasks for the \bench{MeMN2N} model with the \bench{bAbI} dataset.
\sys achieves the average pruning rates of 78.6\% (max. 93.2\%) and 75.5\% (max. 93.0\%) for the \bench{BERT-B} and \bench{BERT-L} models with the \bench{GLUE} dataset, while achieving 73.9\% and 74.1\% with the {SQUAD} dataset, respectively. 
Moreover, \sys provides a 72.6\% pruning rate for \bench{ALBERT-XX-L} with the \bench{SQUAD} dataset, 60.3\% for \bench{ViT-B} with the \bench{CIFAR-10} dataset, and 73.9\%  for \bench{GPT-2-L} with the \bench{WikiText-2} dataset.
As the results suggest, \sys can significantly prune out the $Score$s across various tasks, with greater benefits to \bench{MeMN2N} tasks compared to the \bench{BERT} ones.
We conjecture the lower pruning rates in \bench{BERT} models are due to the higher probability of correlation between various tokens in the more complex language processing tasks compared to \bench{MemN2N}.

\begin{table*}
\tiny
\footnotesize{
\centering
\caption{\label{table:comparison}\sys performance comparison under different scenarios with prior work~\cite{a3:hpca20,spatten:hpca21}.}
\resizebox{0.99\textwidth}{!}{
\begin{tabular}{l|c|c|c|c|c|c|c|c}
\bottomrule
\bftable{Metric~(unit)}&\bftable{A$^\mathrm{3}$-Base}&\bftable{A$^\mathrm{3}$-Conserv}&\bftable{SpAtten}&\bftable{HP-\sys}&\bftable{HP-\sys$^\dagger$}&\bftable{HP-\sys$^\ddagger$}&\bftable{HP-\sys$^{\dagger*}$}&\bftable{HP-\sys$^{\ddagger*}$}\\\bottomrule
\bftable{Process~(nm)}&\normtable{40}&\normtable{40}&\normtable{40}&\normtable{65}&\normtable{40}&\normtable{40}&\normtable{40}&\normtable{40}\\\midrule
\bftable{Area~(mm$^2$)}&\normtable{2.08}&\normtable{2.08}&\normtable{1.55}&\normtable{3.47}&\normtable{1.31}&\normtable{1.31}&\normtable{1.05}&\normtable{1.05}\\\midrule
\bftable{Key Buffer~(KB)}&\normtable{20}&\normtable{20}&\normtable{24}&\normtable{48}&\normtable{24}&\normtable{24}&\normtable{24}&\normtable{24}\\\midrule
\bftable{Value Buffer~(KB)}&\normtable{20}&\normtable{20}&\normtable{24}&\normtable{64}&\normtable{24}&\normtable{24}&\normtable{24}&\normtable{24}\\\midrule
\bftable{($\mathcal{Q}$, $\mathcal{K}$)-bits}&\normtable{(9, 9)}&\normtable{(9, 9)}&\normtable{(12, 12)}&\normtable{(12, 12)}&\normtable{(12, 12)}&\normtable{(12, 12)}&\normtable{(9, 9)}&\normtable{(9, 9)}\\\midrule
\bftable{GOPs~/~s}&\normtable{259.0}&\normtable{518.0}&\normtable{728.4}&\normtable{574.1}&\normtable{932.8}&\normtable{1084.9}&\normtable{1143.9}&\normtable{1330.3}\\\midrule
\bftable{GOPs~/~J}&\normtable{2354.5}&\normtable{4709.1}&\normtable{772.9}&\normtable{519.3}&\normtable{2224.8}&\normtable{2028.8}&\normtable{3353.8}&\normtable{3058.4}\\\midrule
\bftable{GOPs~/~s~/~mm$^2$}&\normtable{124.5}&\normtable{249.0}&\normtable{470.0}&\normtable{165.5}&\normtable{710.4}&\normtable{826.1}&\normtable{1093.8}&\normtable{1272.1}\\\bottomrule
\end{tabular}
}\\\vspace{0.1cm}
\footnotesize{$\dagger$ Dennard scaling trend applied to map on 40 nm process -- \rev{$\ddagger$ Scaling rule from~\cite{scale_measure} applied to map on 40 nm process} -- *scaled to 9 bit $\mathcal{Q}$, $\mathcal{K}$}}
\end{table*}
As Figure~\ref{fig:sparsity} shows, in the case of \bench{ALBERT-XX-L} with \bench{SQUAD}, we see more pruning opportunities compared to \bench{BERT}, presumably because of its larger model architecture with more redundant computations.
Similar trend is observed for \bench{GPT-2-L}.
With regard to \bench{ViT-B}, we see lower pruning compared to NLP tasks, commensurate with prior studies~\cite{chasingvision:2021}.
This occurs because information is more local in images compared to texts, and therefore there is less redundancy in the attention layers for vision tasks.

\begin{figure}
\centering
\includegraphics[width=0.85\columnwidth]{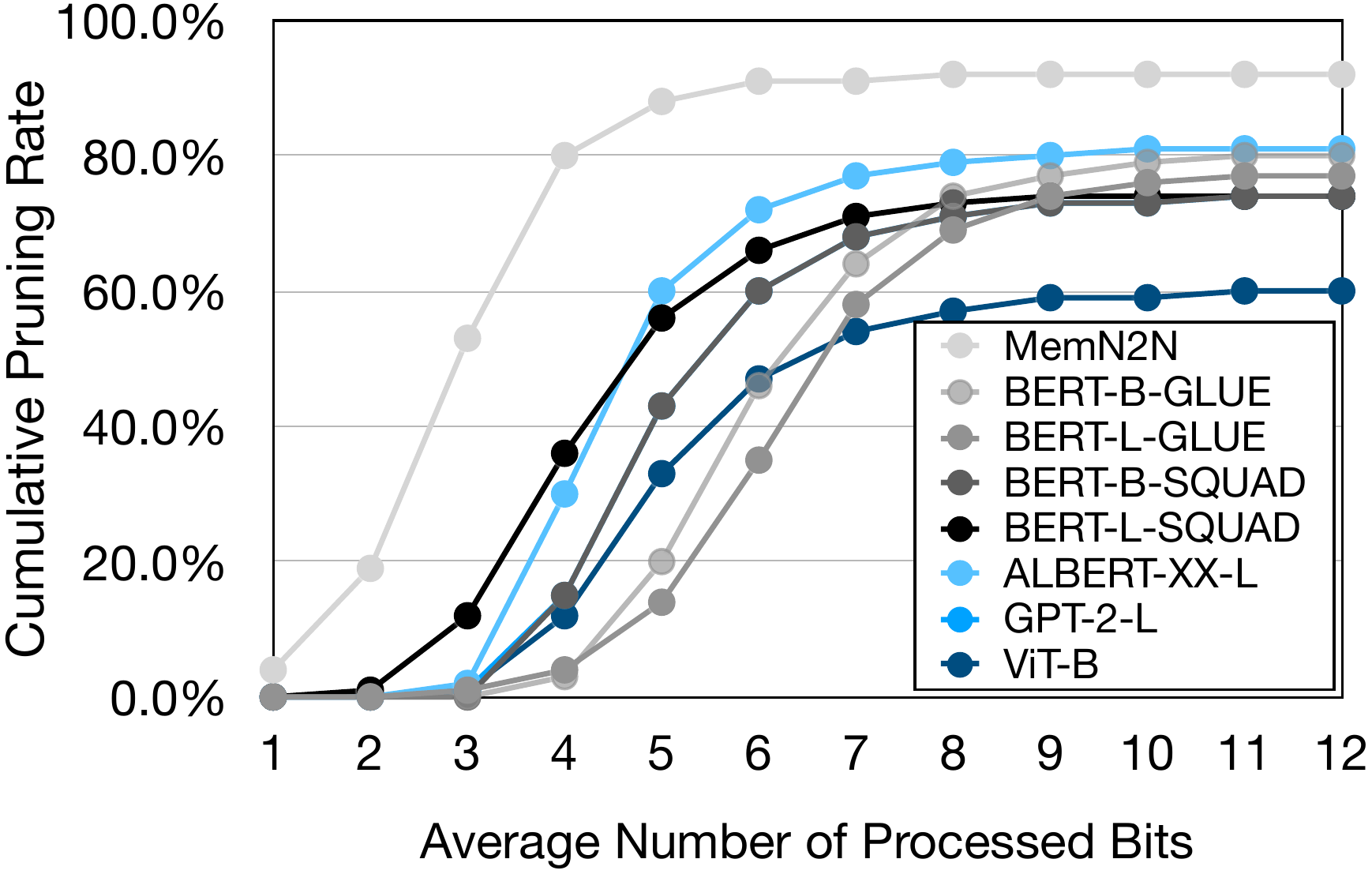}
\caption{\rev{Cumulative pruning rate with respect to the number of bits processed during bit-serial early termination. Each line obtained by averaging across all the pruning rates per task.}} 
\label{fig:bit_eary_stop}
\end{figure}

\niparagraph{Bit-level early-compute termination.}
Figure~\ref{fig:bit_eary_stop} depicts the proposed bit-level early compute termination feature and its relation with the achieved runtime pruning rates.
The x-axis shows the number of bits processed sequentially,  while the y-axis shows the cumulative achieved pruning rate averaged over all of the datasets' tasks.
Intuitively, as more bits are processed during $Score$ computations, the dynamic margin becomes smaller and thus the pruning rate increases.
As shown, as the average number of processed bits increases, the cumulative pruning rate gradually plateaus, indicating saturation. In this scenario, the higher number of bits are only required for fully calculating unpruned $Score$s.
We establish that the lower redundancy in model parameters of some transformer models, e.g. \bench{BERT-L}~/~\bench{ViT-B}, hinders higher runtime pruning.
Because lower redundancy generally translates to a higher number of average bits calculations, it proportionally diminishes the potential gains from bit-wise early termination.
Averaged over pruned $Score$s in bit-serial mode, \bench{MemN2N} with the \bench{bAbi} dataset requires 4.5 bits, while \bench{BERT-B} and \bench{BERT-L} require 8.3 and 8.0 bits with the \bench{GLUE} dataset.
With the \bench{SQUAD} dataset, the average number of bits  in \bench{BERT-B} and \bench{BERT-L} are 7.6 and 9.0 bits, whereas \bench{ALBERT-XX-L} maintains 8.0 bits.
The average number of bits in \bench{GPT-2-L} and \bench{ViT} attain 7.6 bits and 8.5 bits, respectively.
This devised early-termination mechanism significantly reduces the computations of the $\mathcal{Q} \times \mathcal{K}\rev{^T}$.
\subsection{Accelerator Performance Results} 
\label{subsec:perf_results}
\begin{figure*}
\centering
\includegraphics[width=1.96\columnwidth]{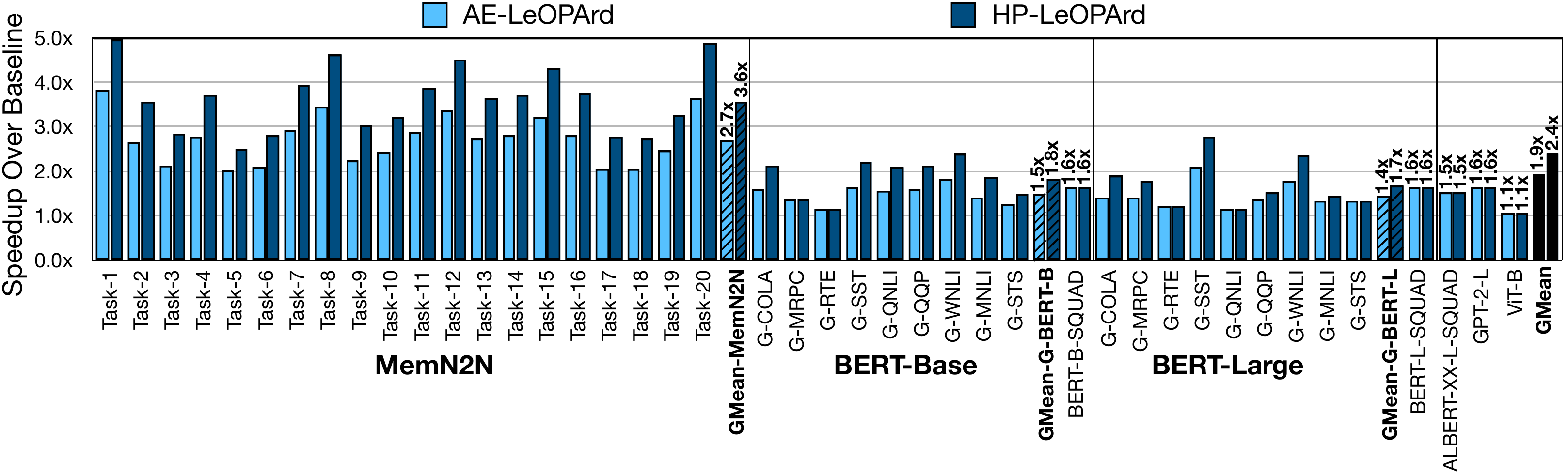}
\caption{\rev{Speedup comparison to baseline design for \aeleopard and \hpleopard (prefix "G-": \bench{GLUE} dataset).}}
\label{fig:speedup} 
\end{figure*}

\niparagraph{Performance and energy comparison to baseline.}
Figure~\ref{fig:speedup} shows the speedup improvements delivered by \sys compared to the baseline design, across all the \rev{43} studied tasks.
In this comparison, we consider the total execution runtime for all attention layers of the models.
On average across all tasks, \aeleopard and \hpleopard provide \rev{\SpeedupOverBaselineAE} and \rev{\SpeedupOverBaselineHP} speedup over the baseline, respectively.
These improvements stem from both \sys runtime pruning that reduces operations on the back-end unit (e.g., Softmax and $\times \mathcal{V}$) and bit-level early compute termination that saves cycles on $\mathcal{Q} \times \mathcal{K}\rev{^T}$ computations for pruned $Score$s.
Across the workloads, \sys delivers higher speedups for \bench{MemN2N} compared to the other benchmarks. We attribute these improvements to the higher pruning rate and consequently more bit-level termination opportunities in this model's tasks.
Among all the tasks, \bench{MemN2N-Task-1} enjoys the maximal speedup (3.8$\times$ for \aeleopard and 5.1$\times$ for \hpleopard) while \bench{ViT-B} gains the minimal improvements (1.1$\times$ for both \aeleopard and \hpleopard).
The benefits are more pronounced for \hpleopard because it deploys more \code{QK-DPU}s, which both improves the performance of the front-end \code{Q-PU} unit, and delivers more inputs ($Score$s) to the back-end stage. 
The latter generally increases the back-end utilization.
\begin{figure*}
\centering
\includegraphics[width=1.97\columnwidth]{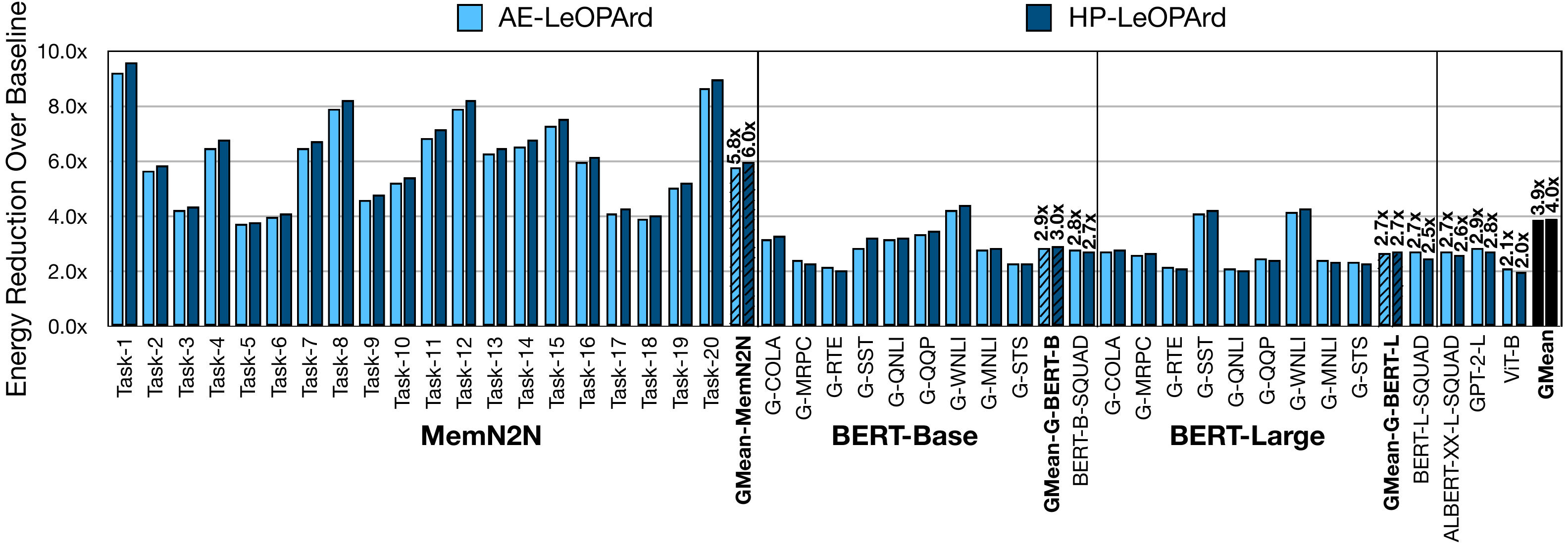}
\caption{\rev{Total energy reduction for \aeleopard and \hpleopard compared to baseline (prefix "G-": \bench{GLUE} dataset).}}
\label{fig:energy} 
\end{figure*}

Figure~\ref{fig:energy} compares the energy reduction (including compute and on-chip memory accesses) achieved by \sys to the baseline.
On average, \sys reduces total energy consumption by a factor of \rev{\EnergyOverBaselineAE} for \aeleopard and \rev{\EnergyOverBaselineHP} for \hpleopard, across all the studied tasks.
Similarly to the speedup comparisons, \bench{MemN2N} enjoys a greater energy reduction than the other benchmarks due to the higher pruning rate and therefore faster bit-level compute terminations.
Across all tasks, the energy reduction is the greatest for \bench{MemN2N-Task-1} (9.2$\times$ for \aeleopard and 9.6$\times$ for \hpleopard) and \bench{ViT-B} achieves the lowest savings ($\approx$\,2.0$\times$ for \aeleopard and \hpleopard).
The impact of \sys on energy exceeds that on speedup, because runtime pruning and bit-level early termination reduce computation energy (contributing to both energy savings and speedup) and memory accesses (only contributing to energy savings).
The energy reductions for both \aeleopard and \hpleopard are not substantially different.
Because the additional \code{QK-DPU}s in \hpleopard increase both power and performance, total energy consumption remains similar. 

\begin{figure}
\centering
\includegraphics[width=0.99\columnwidth]{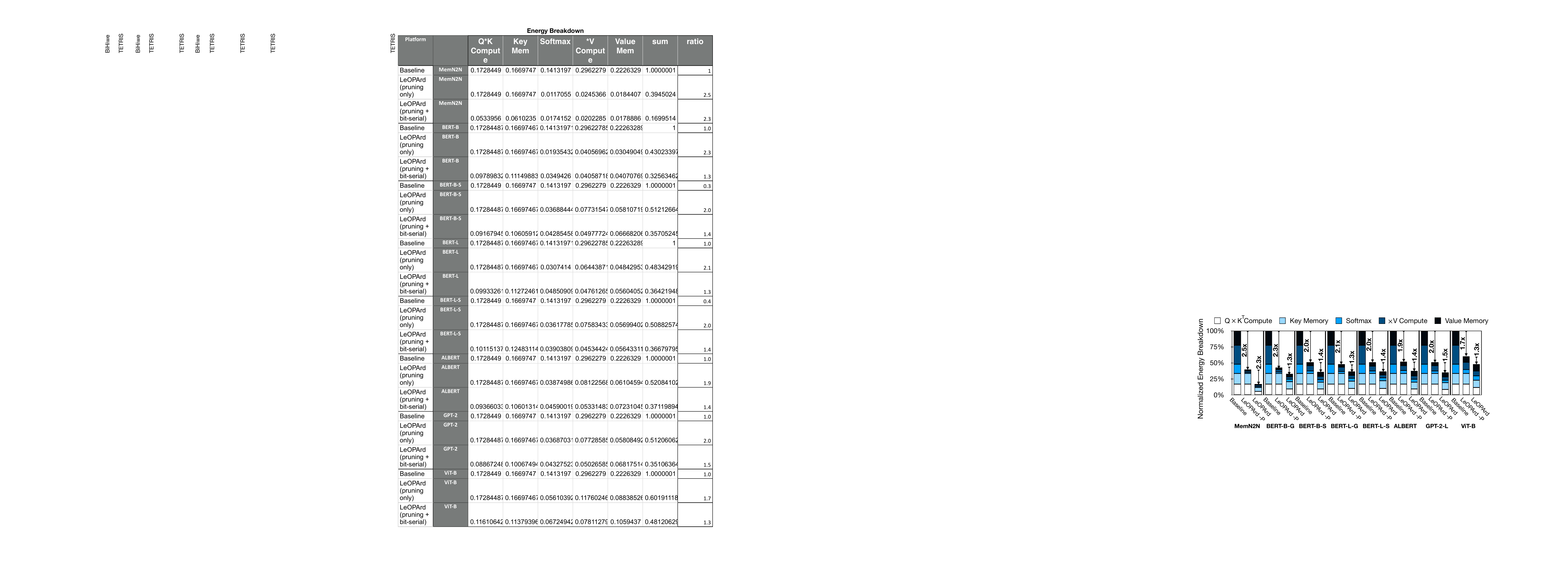}
\caption{Normalized \leopard's average energy breakdown and the contribution of runtime pruning and bit-level early termination in energy saving (\sys-P: with only pruning, and \sys: pruning $+$ bit-serial early termination) across one transformer head.}
\label{fig:energy_saving_contribution}
\end{figure}

\niparagraph{Analysis of energy savings breakdown.}
Figure~\ref{fig:energy_saving_contribution} analyzes the breakdown of total energy consumption across five microarchitectural components: (1) $\mathcal{Q} \times \mathcal{K}\rev{^T}$ computations, (2) $\mathcal{K}$ buffer memory access, (3) Softmax, (4) $\times \mathcal{V}$ computations, and (5) value buffer memory access.
We report the average breakdown across all tasks for each workload.
Additionally, Figure~\ref{fig:energy_saving_contribution} illustrates the contribution of \sys's two main optimizations: (1) runtime pruning and (2) early compute termination through bit-serial execution to the overall energy savings in \aeleopard.
We normalize the energy breakdowns to a baseline, which does not utilize any of the \sys's optimizations. 
In the baseline, $\times \mathcal{V}$ computations and value buffer memory accesses proportionally consume the highest energy due to the lack of runtime pruning; ergo, higher average number of bits in $\mathcal{Q}\times\mathcal{K}^T$.
Recall that the \sys's back-end unit encloses Softmax, $\times \mathcal{V}$, and its associated buffer accesses.
As the results show, this unit consumes more than 65$\%$ of the total energy in the baseline design.
\sys's runtime pruning enables skipping computations and memory accesses for inconsequential $Score$s during the back-end processing, delivering 1.7$\times$ (\bench{ViT-B}) to 2.5$\times$ (\bench{MemN2N}) energy savings.
For these tasks, the bit-serial execution in \sys along with its early termination brings further energy savings of 1.3$\times$ (\bench{ViT-B}) to 2.3$\times$ (\bench{MemN2N}) on top of runtime pruning.
These additional benefits arise from avoiding the inconsequential bit  computations in $\mathcal{Q} \times \mathcal{K}$ and their associated $\mathcal{K}$ buffer accesses.

\niparagraph{Comparison with \aaa and SpAtten.}
~Table~\ref{table:comparison} compares the characteristics and performance of \hpleopard and its scaled versions with \aaa and SpAtten.
Compared to SpAtten, \hpleopardd\rev{(\hpleoparddd)} delivers 3$\times$ \rev{(2.6$\times$)} improvements in GOPs~/~J and 1.5$\times$ \rev{(1.7 $\times$)} improvements in GOPs~/~\rev{s}~/~mm$^2$, while both designs have virtually no model accuracy degradation.
These benefits are attributed to the \sys's higher pruning rate and to the bit-level early compute termination.
For comparison with \aaa, we evaluate \hpleopardds \rev{(\hpleoparddds)}, which are scaled to 40~nm and deploy 9-bit arithmetic for $\mathcal{Q} \times \mathcal{K}\rev{^T}$. A$^\mathrm{3}$-Conservative deploys heuristic approximation to minimize accuracy degradation on top of A$^\mathrm{3}$-Base, which does not use approximation.
\hpleopardds (\hpleoparddds) achieves 1.4$\times$ \rev{(1.3$\times$)} higher energy efficiency (in GOPs~/~J) and 8.8$\times$ (10.2$\times$) area efficiency (in GOPs~/~\rev{s}~/~mm$^2$) than A$^\mathrm{3}$-base.
\hpleopardds (\hpleoparddds) also provides 4.4$\times$ (5.1$\times$) improvements in terms of GOPs~/~s~/~mm$^2$ compared to A$^\mathrm{3}$-Conservative.
Although A$^\mathrm{3}$-Conservative provides 29\% and 35\% higher energy efficiency compared to \hpleopardds and \hpleoparddds, respectively, this comes at the cost of visible accuracy degradation, e.g., 1.0\% for \bench{MemN2N} and 1.3\% for \bench{BERT-Base} with the \bench{SQUAD} dataset as reported in \cite{a3:hpca20}.
On the other hand, \sys's carefully crafted  gradient-based training balances pruning rate and model accuracy, providing accuracy degradation of only 0.06\% and 0.26\% for the aforementioned models and datasets without manual configurations for heuristic parameters.

\begin{figure}[t]
\centering
\subfloat[]{\includegraphics[width=0.53\columnwidth]{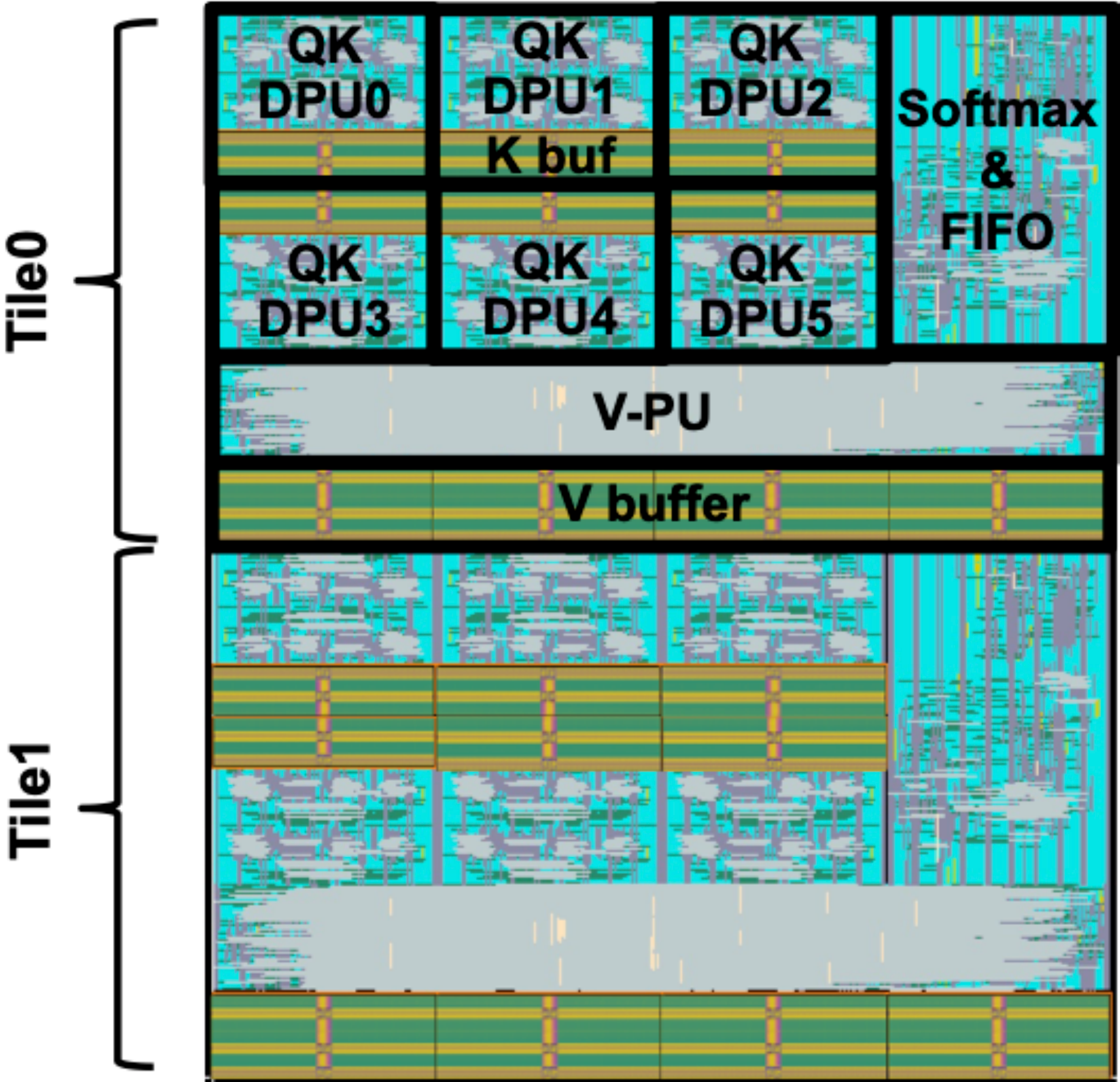}}
\hspace{0.4cm}
\subfloat[]{\includegraphics[width=0.35\columnwidth]{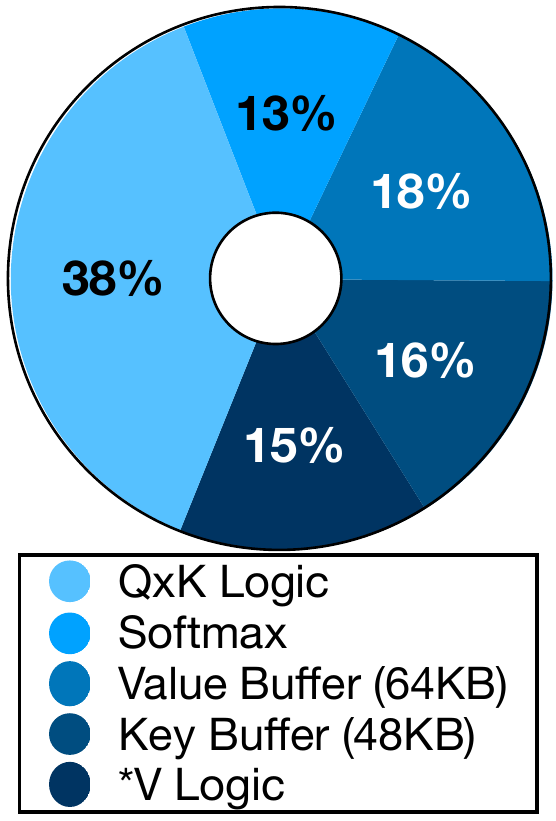}}
\caption{AE-\sys: (a) layout ($2.3\times 2.8~\mathrm{mm^2}$) and (b) area breakdown.}
\label{fig:layout}
\end{figure}

\niparagraph{\sys accelerator layout area details.}
Figure~\ref{fig:layout}(a) shows the layout of \sys architecture, which occupies $2.3\times2.8~ \mathrm{mm^2}$, including two tiles.
The layouts are generated by meeting the design rule check in a 65~nm process and targeting 65-75\% physical density, commonly used for the routing convenience and tape-out yield.
Figure~\ref{fig:layout}-(b) reports the area breakdown, where \code{QK-DPU} takes the largest proportion as we employ $N_{\mathrm{QK}}$ \code{QK-DPU} in consideration of the high pruning rate. 
This leads to 56\% area occupied by the front-end unit, which includes \code{QK-DPU} and $\mathcal{K}$ buffer.
The on-chip memory for $\mathcal{K}$ and $\mathcal{V}$ occupies 34\% of the layout area.
\subsection{Architecture Design Space Exploration} \label{sec:design_space}
\niparagraph{\code{QK-PU} parallelism degree.}
\label{subsec:QK processor parallelism}
As discussed in Section~\ref{sec:overall_arch}, the number of \code{QK-DPU}s ($N_{\mathrm{QK}}$) within one \code{QK-PU} exhibits a trade-off space in designing the \sys accelerator.
\begin{figure}
\centering
\includegraphics[width=0.88\columnwidth]{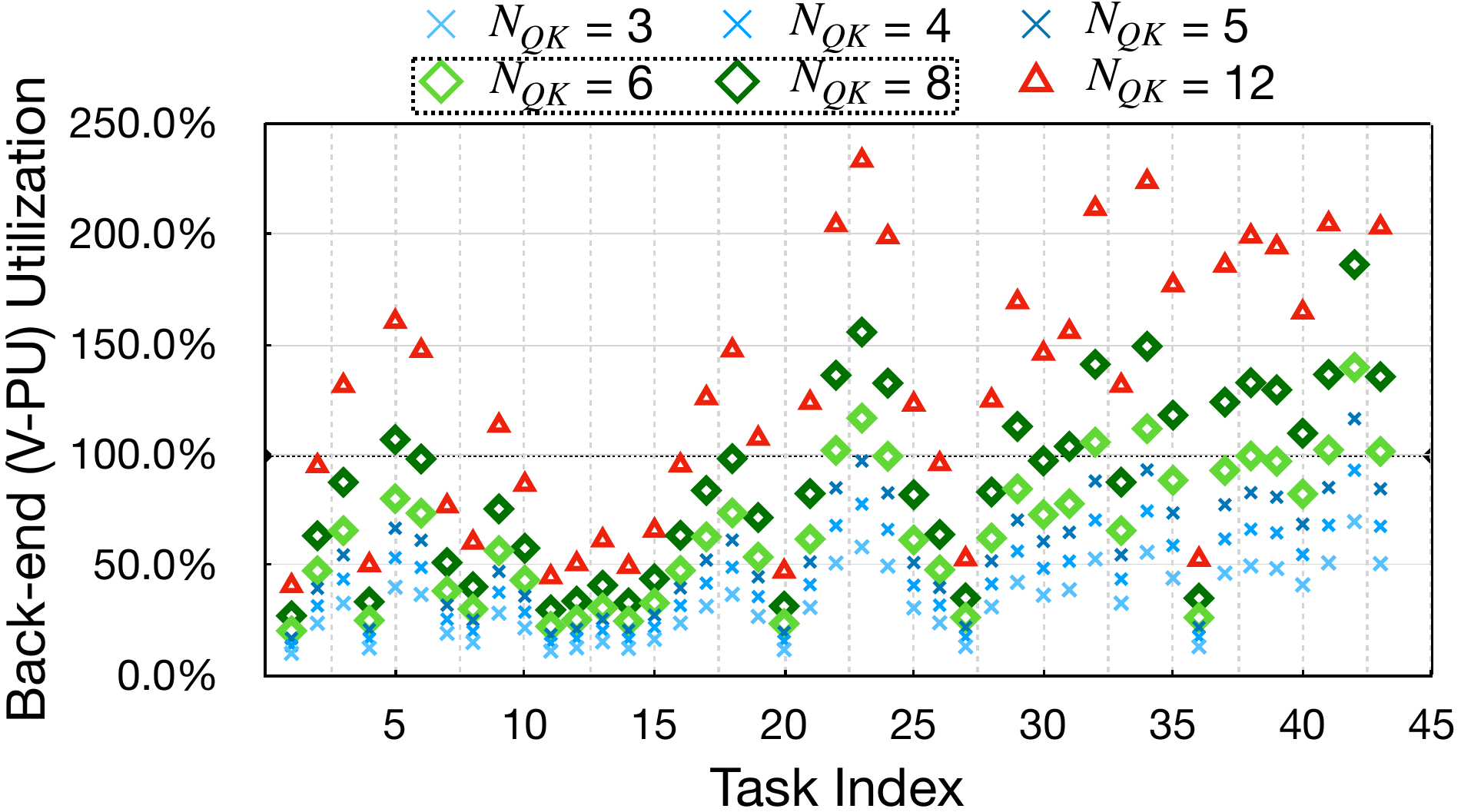}
\caption{Back-end \code{V-PU} utilization over the \code{QK-PU} parallelism ($N_{\mathrm{QK}}$). $N_{QK}=6$ and $N_{QK}=8$ form the favorable configurations in terms of back-end utilization in \aeleopard and \hpleopard, respectively.}
\label{fig:backend_utilization}
\end{figure}
To find the number of \code{QK-DPU}s that balances the utilization of front-end and back-end units, we sweep the $N_{\mathrm{QK}}$ from three to 12 in Figure~\ref{fig:backend_utilization} and report the \code{V-PU} utilization across the evaluated tasks.
If utilization exceeds 100$\%$ (common when $N_{\mathrm{QK}} = 12$), the back-end \code{V-PU} is over-subscribed due to the throughput mismatch between \code{V-PU} and \code{QK-PU}.
This mismatch throttles the back-end \code{V-PU} and turns into the system bottleneck, frequently stalling the front-end.
On the other hand, when $N_{\mathrm{QK}}=3$, the \code{V-PU} is chronically under-utilized due to a significant reduction in its number of computations, attributed to front-end runtime pruning mechanism.
As marked by dark green diamonds, $N_{\mathrm{QK}}=8$ adequately balances the \code{V-PU} utilization and the number of front-end unit stalls.
Thus, we favor this configuration for \hpleopard.
The second best configuration to balance front- and back-end utilization is $N_{\mathrm{QK}}=6$ (marked by light green diamonds).
As such, we choose this configuration for \aeleopard, which matches the baseline chip area usage.

\begin{figure}
\centering
\includegraphics[width=0.77\columnwidth]{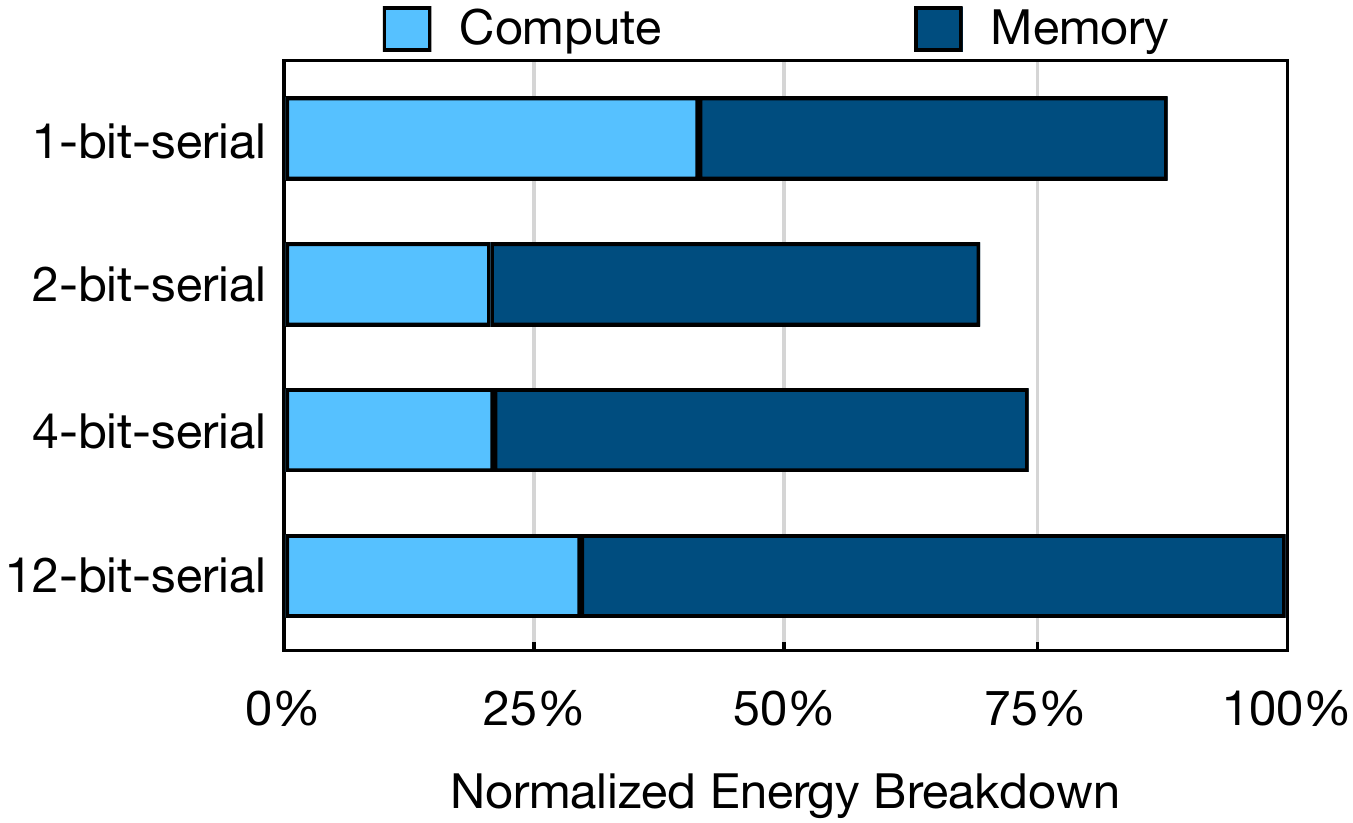}
\caption{Design space exploration for the resolution ($\mathcal{B}$) of bit-serial execution with respect to normalized average \code{QK-DPU} energy per $Score$.}
\label{fig:bit_vs_power}
\end{figure}

\niparagraph{Bit-serial processing granularity.}
~Figure~\ref{fig:bit_vs_power} illustrates the design space exploration for granularity of the bit-serial execution in \code{QK-DPU} ($\mathcal{B}$).
This bit-level granularity creates a trade-off space, where decreasing the $\mathcal{B}$ stores intermediate results at the end of each bit processing cycle more frequently (escalating the energy).
At the same time, increasing $\mathcal{B}$ curtails the performance of early compute termination due to lower resolution in stopping the computations.
To find the optimal point, we sweep the $\mathcal{B}$ for values of 1, 2, 4, and 12 bits and measure the average consumed energy and its breakdown ($\mathcal{Q} \times \mathcal{K}\rev{^T}$ logic and key buffer accesses) per one output $Score$.
All the numbers are normalized to 12-bit processing that does not employ any bit-serial execution.
Figure~\ref{fig:bit_vs_power} depicts this analysis for \bench{MemN2N} tasks (results for \rev{other} models are similar) and reports the average across all tasks.
As shown, 2-bit-serial execution strikes the right balance between energy consumption of the bit-serial computations and the resolution of bit-level early compute termination.
\section{Related Work}
\label{sec:related}
In contrast with prior work, \sys explores a distinct design space for accelerating attention models through gradient-based learned runtime pruning.
This tight integration of pruning and training enables \sys to reduce the computation cost with virtually zero accuracy degradation across a range of language and vision transformer models.
Building on these algorithmic insights, we devise a bit-serial execution strategy that conservatively terminates the computations as early as possible. 
Below, we cover the most relevant work and position our paper with respect to it.

\niparagraph{Hardware-algorithm co-design for attention models.}
Several algorithmic optimizations co-designed with hardware acceleration were proposed for efficient execution of attention models~\cite{elsa:isca21,spatten:hpca21,sanger:micro21,edgebert:micro21,softermax:dac21,a3:hpca20,optimus:mlsys20,gobo:micro20}.
\aaa has proposed an \emph{approximation} method with a hardware accelerator to prune out the ineffectual computations in attention.
This method searches effective data during the $query\times key$ operation in addition to another approximation mechanism after score calculation.
SpAtten~\cite{spatten:hpca21} prunes the ineffectual input tokens and heads, in addition to progressive quantization during computations at runtime, to improve the performance and memory bandwidth.
We provide a head-to-head comparison to these works in Section~\ref{subsec:perf_results}.
ELSA~\cite{elsa:isca21} aims to address the costly candidate search process of $A^3$ and incorporates a user-defined "confidence-level" parameter to find the optimal thresholds from training statistics.
EdgeBERT~\cite{edgebert:micro21} leverages entropy-based early exiting technique to predict the minimal number of transformer layers that need to be executed, while the rest can be skipped.
Other works aim to address the computational cost of self-attention via sparse matrix operation~\cite{optimus:mlsys20, sanger:micro21,chen2021chasing}, quantization~\cite{gobo:micro20}, and Softmax approximation~\cite{softermax:dac21}. 
Moreover, none of these prior designs explored \emph{bit-level} early compute termination.

\niparagraph{Algorithmic optimizations for transformer acceleration.}
Another line of prior inquiry proposes only algorithmic optimizations to provide sparsity in computing attention models.
Proposals in~\cite{rounting-transformer,longformer:2020,ye2021tr,reformer:2020,blockwise,bp-transformer, child2019generating,michel2019sixteen,wang2019structured,wen2017learning} offer static sparsity in the attention layers to reduce its significant computational cost. 
Other work~\cite{explicit, adaptively-sparse, fine-tune-bert} provides dynamic sparsity based on the input samples, yet still requires full computation of the $\mathcal{Q} \mathcal{\times} \mathcal{K}\rev{^T}$.
Our proposal fundamentally differs from this prior seminal work, because it formulates the problem of pruning threshold finding as a regularizer to methodically co-optimize with the weight parameters of the models, \emph{without approximation}.
Additionally, \sys provides architectural support to stop the attention computations as early as possible during runtime.

\niparagraph{Early compute-termination in DNNs.}
Prior work~\cite{zap:eccv20,snapea:isca:2018,compend:ics18,predictivenet:iscas17} has proposed techniques to early terminate the computations of convolution layers by leveraging the zero production feature of ReLU for negative numbers.
In contrast, this work focuses on early termination of a fundamentally different operator, attention in transformers, and provides unique mechanisms to enable that.
Moreover, the prior works consider zero as a fixed threshold in their methods, but \sys formulates the thresholds as a regularizer and \emph{finds} layer-wise values through gradient descent optimization to preserve the accuracy of the models.

\niparagraph{DNN acceleration.}
A large swath of work~\cite{yazdanbakhsh2021evaluation,bihiwe:pact20,bpvec:dac20,sigma:hpca20,planaria:micro20,bit-tactical:asplos19,encodeep,bitblade:dac19,laconic:isca19,fastwave:iccad19,tangram:asplos19,eyerissv2:journal19,simba:micro19,maestro:micro19,maeri:asplos18,brainwave:isac:2018,bitfusion:isca18,flexigan:fccm18,deepfense:iccad18,unpu:isscc:2018,ganax:isca:2018,loom:dac18,neuralcache:isca18,scnn:isca:2017,bit-pragmatic:micro17,pipelayer:hpca:2017,tetris:asplos:2017,tpu:isca:2017,eie:isca:2016,stripes:micro:2016,minerva:isca:2016,neurocube:isca:2016,isaac:isca:2016,prime:isca:2016,cambricon-x:micro:2016,cnvlutin:isca:2016,cambricon:isca:2016,eyeriss:isca:2016,dnnweaver:micro:2016,yazdanbakhsh2015neural,anpu,dadiannao:micro:2014} is dedicated to accelerating DNNs.
Although inspiring, these designs do not deal with the challenges unique to the attention mechanisms of transformers, as opposed to this work.
\section{Conclusion}
\label{sec:conclusion}
Transformers through the self-attention mechanism have triggered an exciting new wave in machine learning, notably in Natural Language Processing (NLP).
The self-attention mechanism computes pairwise correlations among all the words in a subtext.
This task is both compute and memory intensive and has become one of the key challenges in realizing the full potential of attention models. 
One opportunity to slash the overheads of the self-attention mechanism is to limit the correlation computations to a few high score words and computationally prune the inconsequential scores at runtime through a thresholding mechanism.
This work exclusively formulated the threshold finding as a gradient-based optimization problem.
This formulation strikes a formal and analytical balance between model accuracy and computation reduction.
To maximize the performance gains from thresholding, this paper also devised a bit-serial architecture to enable an early-termination atop pruning with no repercussions to model accuracy.
These techniques synergistically yield significant benefits both in terms of speedup and energy savings across various transformer-based models on a range of NLP and vision tasks.
The application of the proposed mathematical formulation of identifying threshold values and its cohesive integration into the training loss is broad and can potentially be adopted across a wide range of compute reduction techniques. 
\begin{acks}
Soroush Ghodrati is partly supported by a Google PhD Fellowship.
This work was in part supported by generous gifts from Google, Samsung, Qualcomm, Microsoft, Xilinx as well as the National Science Foundation (NSF) awards CCF\#2107598, CNS\#1822273, National Institute of Health (NIH) award \#R01EB028350, Defense Advanced Research Project Agency (DARPA) under agreement number \#HR0011-18-C-0020, and Semiconductor Research Corporation (SRC) award \#2021-AH-3039. The U.S. Government is authorized to reproduce and distribute reprints for Governmental purposes not withstanding any copyright notation thereon. The views and conclusions contained herein are those of the authors and should not be interpreted as necessarily representing the official policies or endorsements, either expressed or implied of Google, Qualcomm, Microsoft, Xilinx, Samsung, NSF, SRC, NIH, DARPA or the U.S. Government.
We also would like to extend our gratitude towards Cliff Young, Suvinay Subramanian, Yanqi Zhou, James Laudon, and Stella Aslibekyan for their invaluable feedback and comments.
\end{acks}

\bibliographystyle{ACM-Reference-Format}
\bibliography{main}
\end{document}